\documentclass[a4paper,fleqn]{cas-dc}
\usepackage[numbers]{natbib}
\usepackage{amsmath}

\def\tsc#1{\csdef{#1}{\textsc{\lowercase{#1}}\xspace}}
\tsc{WGM}
\tsc{QE}

\begin{document}
\let\WriteBookmarks\relax
\def\floatpagepagefraction{1}
\def\textpagefraction{.001}
\shorttitle{}    

\shortauthors{}

\title [mode = title]{A Cross-Hierarchical Difference Feature Fusion Network Based on Multiscale Encoder–Decoder for Hyperspectral Change Detection}                      

\author[a]{Mingshuai Sheng}[orcid=0000-0001-8860-7250]
\credit{Writing – original draft, Software, Conceptualization, Data curation}
\ead{mingshuai@hainanu.edu.cn}
\affiliation[a]{organization={School of Information and Communication
		Engineering, Hainan University},
	    city={Haikou},
	    postcode={570228}, 
	    state={Hainan},
	    country={China}}

\author[b]{Uzair Aslam Bhatti\corref{1}}
\credit{Writing – review\& editing, Supervision, Methodology}
\cormark[1]
\ead{uzairaslambhatti@hotmail.com}

\author[b]{Junfeng Zhang}
\ead{jfzhang@hainanu.edu.cn}
\credit{Resources}

\author[b]{Siling Feng}
\ead{fengsiling@hainanu.edu.cn}
\credit{Resources, Methodology}
\affiliation[b]{organization={Department of Artificial Intelligence, School of Information and
				Communication Engineering, Hainan University},
		city={Haikou},
		postcode={570228}, 
		state={Hainan},
		country={China}}

\author[c]{Yonis Gulzar}
\ead{younisgulzar@gmail.com}
\credit{Resources}
\affiliation[c]{organization={Department of Management Information Systems, 
		King Faisal University},
	    state={Al-Ahsa},
	    country={Saudi Arabia}}

\cortext[1]{Corresponding author.} 


\begin{abstract}
Hyperspectral change detection (HCD) is one of the core applications of remote sensing images, holding significant research value in fields like environmental monitoring and disaster assessment. However, existing methods often suffer from incomplete capture of multiscale spatial-spectral features and insufficient fusion of differential feature information. To address these challenges, this paper proposes a Cross-Hierarchical Differential Feature Fusion Network (CHDFFN) based on a multiscale encoder-decoder. Firstly, a multiscale feature extraction subnetwork is designed, taking the customized encoder-decoder as the backbone, combined with residual connections and the proposed dual-core channel-spatial attention module to achieve multi-level extraction and initial integration of spatial-spectral features. The encoder embeds convolutional blocks with different receptive field sizes to capture multiscale representations from shallow details to deep semantics. The decoder fuses the encoder’s output via skip connections to gradually restore spatial resolution while suppressing background noise and redundancy. To enhance the model’s ability to capture differential features between bi-temporal hyperspectral images, a spatial-spectral change feature learning module is designed to learn hierarchical change representations. Additionally, an adaptive high-level feature fusion module is proposed, dynamically balancing the contribution of hierarchical differential features by adaptively assigning weights, which effectively strengthens the model’s capability to characterize complex change patterns. Finally, experiments on four public hyperspectral datasets show that compared with some state-of-the-art methods, the average maximum improvements of OA, KC, and F1 are 4.61\%, 19.79\%, and 18.90\% respectively, verifying the model’s effectiveness.
\end{abstract}




\maketitle

\section{Introduction}
\label{sec1}
Hyperspectral images (HSI) provide continuous spectral information and integrate spatial and spectral features, enabling simultaneous representation of spatial structures and spectral signatures \cite{hu2022hypernet}. These properties make HSI highly effective for detecting changes \cite{wang2022spectral} between multi-temporal images, with applications in land cover classification and disaster assessment \cite{liu2019review}. Hyperspectral change detection (HCD) has broad potential \cite{hou2021hyperspectral,ding2025survey} in environmental monitoring, disaster loss assessment, and agricultural resource investigation by analyzing data from different time phases to detect changes in land surface types, distribution, and states. This information supports resource management, environmental governance, and disaster response \cite{ou2022hyperspectral}. The core objective of HCD is to identify differences across two temporal instances, with the main challenge being discrimination between changed and unchanged regions \cite{shafique2023ssvit} in high-dimensional spectral space.

HCD techniques are generally classified into two major categories: traditional methods and deep learning-based approaches\cite{lv2025multiscale}. Early studies primarily relied on statistical and algebraic techniques, with representative methods including spectral angle mapping\cite{zhuang2016strategies} (SAM) and change vector analysis\cite{saha2019unsupervised} (CVA). These methods detect spectral variations by computing indicators such as the angle and Euclidean distance between spectral vectors. Another group of traditional methods involves feature transformation and dimensionality reduction, such as principal component analysis\cite{deng2008pca} (PCA), K-means\cite{lv2019novel} clustering, and independent component analysis\cite{zhong2006multi} (ICA), which aim to extract meaningful change features from complex hyperspectral data. Despite their contributions, traditional approaches suffer from several limitations: they are highly sensitive to predefined parameter settings and exhibit poor robustness to perturbations such as spectral noise and illumination variations, making it difficult to ensure reliable detection performance.

In recent years, deep learning (DL) has achieved remarkable success in various fields, including computer vision and remote sensing. Compared with traditional change detection models, DL model offer advantages such as automatic feature learning from data, nonlinear modeling of change patterns, strong generalization capability, and efficient fusion of multi-dimensional information. These advantages have established DL as a dominant approach in current hyperspectral change detection research. To improve the performance of HCD algorithms and overcome the limitations of traditional methods, researchers have proposed various DL-based feature extraction and fusion schemes\cite{zhou2023spectral,song2018change,luo2024dcenet,jiang2025adaptive}, aiming to enhance detection accuracy and more precisely characterize the changed regions in hyperspectral images.

Convolutional neural networks (CNNs), as a classic deep learning architecture, achieve efficient joint extraction of local spatial and spectral features from input data within an end-to-end learning framework. Ou et al.\cite{ou2022cnn} proposed a CNN framework combining slow-fast band selection (SFBS) and feature fusion grouping (SFBS-FFGNET), which effectively extracts and separates changed and unchanged features, thereby improving change detection accuracy. Seydi et al.\cite{seydi2021new} introduced a novel architecture for binary and multi-class hyperspectral change detection based on spectral unmixing and CNN, leveraging pseudo-data generated by the model to train the network for producing multiple change feature maps. Zheng et al.\cite{zheng2021clnet} designed a cross-layer CNN architecture based on UNet, which deeply fuses multiscale spatial features with multi-level contextual information through a custom cross-layer block (CLB). Recently, Transformer-based architectures, known for their strength in modeling sequential temporal data, have been integrated into HCD algorithms to better capture temporal dependencies in hyperspectral data. Wang et al.\cite{wang2022spectral} proposed the Spectral-Spatial Temporal Transformer model, which enhances change detection accuracy by extracting and integrating spectral, spatial, and temporal features. Zhang et al.\cite{zhang2023cast} developed a cascaded spectral-aware Transformer architecture aimed at improving temporal relevance and spatial globality extraction, effectively capturing long-term dependencies and spectral-spatial features. Zhang et al.\cite{zhang2024dual} presented a dual-branch siamese spatial-spectral Transformer attention network, designed to comprehensively extract spectral and sequential attribute features via spatial attention and spectral Transformer modules. Lin et al.\cite{lin2024unsupervised} proposed an unsupervised Transformer-based multivariate detection method, combining compressed change vector analysis and IR-MAD for pseudo-training sample generation. This method utilizes IR-MAD for subtle change detection and employs the Transformer attention mechanism to model pixel-wise differences and similarities, yielding more accurate and robust detection results.

The attention mechanism is a computational strategy inspired by the way human attention allocates focus. As a key component in deep learning\cite{li2025gassm}, its core idea is that information processing selectively emphasizes important features while suppressing less relevant ones, thereby enhancing the model’s capability to capture critical information and improving processing efficiency. This dynamic weighting scheme effectively addresses the issue of feature redundancy caused by the fixed receptive fields\cite{ji2024domain} in traditional convolutional networks. In hyperspectral change detection, various attention-based detection frameworks have been developed to improve accuracy. For instance, Gong et al.\cite{gong2021spectral} proposed a novel spectral-spatial attention network (S2AN) that employs adaptive spectral and spatial attention mechanisms to suppress irrelevant information for change detection. Hu et al.\cite{hu2024globalmind} introduced the global multi-head interactive self-attention change detection network (GlobalMind), designed to capture correlations among surface objects and land cover transformations, facilitating more comprehensive data analysis and accurate detection. Jiang et al.\cite{jiang2025adaptive} presented the adaptive center-focused hybrid attention network (ACFHAN), which adaptively enhances spatial regions and spectral channels most pertinent to change while suppressing noise. Ren et al.\cite{ren2025interactive} developed the interactive supervised dual-mode attention network (ISDANet). In the encoding stage, ISDANet utilizes MobileNetV2 to extract bi-temporal features, aggregates multiscale semantic features via NFAM to enhance temporal representation, and integrates self-attention with cross-attention through IAEM to enable deep interaction between bi-temporal features. During decoding, it re-weights differential features using SAM and incorporates supervisory signals to dynamically fuse multiscale features, thereby improving boundary detection accuracy and sensitivity to subtle changes.

Previous studies have identified two core optimization directions in HCD: efficient feature extraction and the design of multi-stage feature fusion mechanisms. First, it is critical to address the trade-off between network depth and feature representation capacity—shallow architectures often fail to extract representative features, whereas overly deep or complex networks may result in both computational and feature redundancy, potentially leading the model to learn noisy or irrelevant features. Second, most existing deep learning-based HCD methods utilize RNN or LSTM architectures for bi-temporal feature fusion. However, these approaches lack a decoupling mechanism between changed and unchanged components, forcing the model to concurrently learn dynamic changes and static background information, which increases the difficulty of detecting subtle changes. Additionally, conventional fusion strategies do not incorporate adaptive weighting based on feature relevance, thereby limiting their ability to highlight dominant change features and accurately capture fine-grained differences. Notably, multiscale feature representations have shown considerable promise in improving micro-change detection\cite{huang2025mcecf,qu2021multilevel}. Therefore, this study proposes a multiscale learning framework that enhances HCD performance by decoupling changed-unchanged components, employing dynamic feature weighting strategies, and integrating multiscale spatial-spectral information.
\begin{figure*}
	\centering
	\includegraphics[width=.9\textwidth]{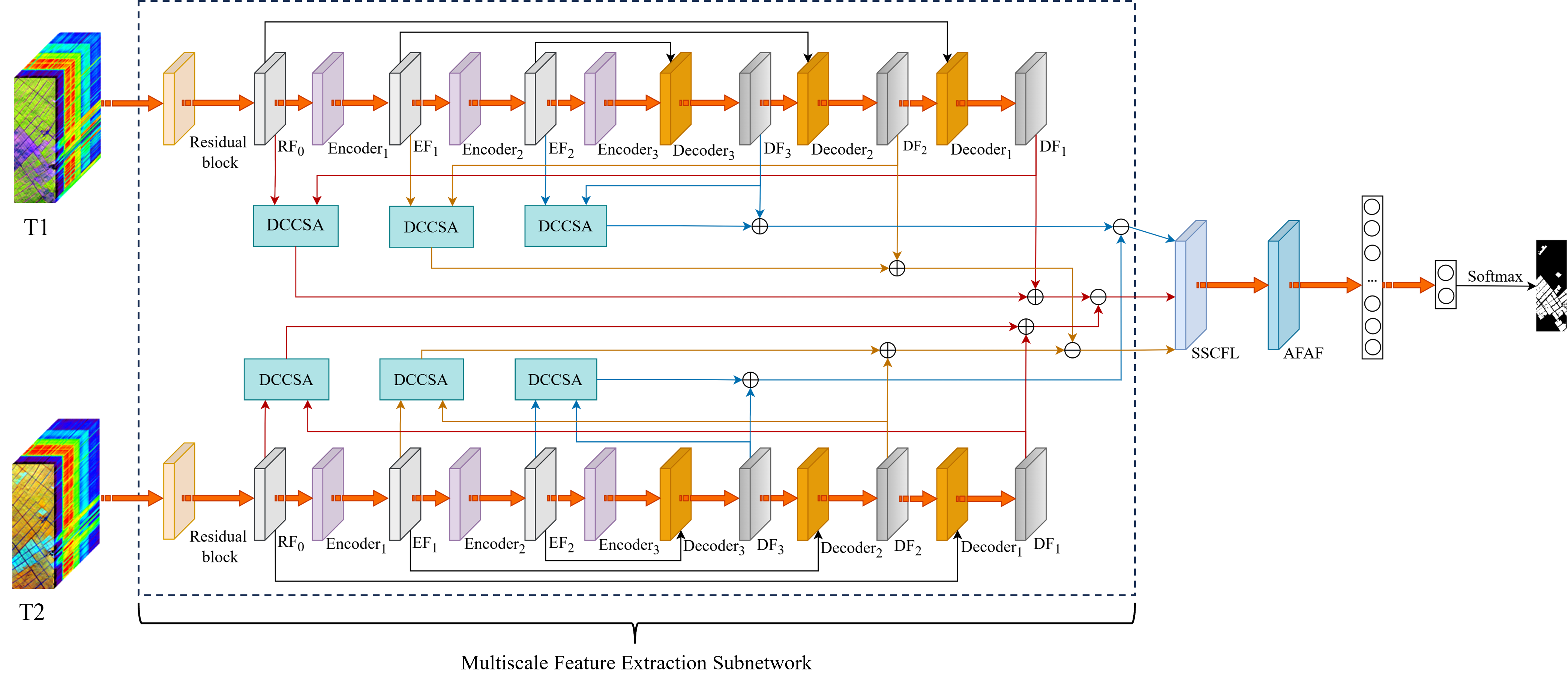}
	\caption{Overall structure diagram of CHDFFN.}\label{fig_1}
\end{figure*}

In summary, this paper proposes a cross-hierarchical difference feature fusion network based on a multiscale encoder–decoder architecture. The proposed CHDFFN consists of four key modules: (1) the dual-core channel-spatial attention module (DCCSA), (2) the multiscale encoder-decoder module (MSED), (3) the spatial-spectral change feature learning module (SSCFL), and (4) the advanced feature adaptive fusion module (AFAF). Among these, the DCCSA and MSED jointly form the spatial-spectral multiscale feature extraction subnetwork. Specifically, DCCSA is designed to identify salient feature channels and emphasize informative spatial regions, while MSED captures hierarchical multiscale representations from hyperspectral data. Subsequently, SSCFL is employed to jointly model spectral, spatial, and temporal dependencies, and AFAF adaptively fuses the extracted features to generate discriminative and change-sensitive representations. The main contributions of this work are summarized as follows:
\begin{itemize}
	\item A DCCSA module is proposed to dynamically allocate attention weights between spectral channels and spatial regions. By emphasizing task-relevant features through a hierarchical attention mechanism, this module enhances the model's ability to focus on salient information.
	
	\item A spatial-spectral multiscale feature extraction subnetwork is designed to extract discriminative spatial-spectral features from bi-temporal hyperspectral images. The encoder-decoder architecture is adopted as the backbone, in which a multiscale convolution module is integrated to expand the receptive field. Furthermore, residual connections are introduced and a dual-core channel-spatial attention module is embedded to effectively fuse contextual information and enhance feature representation.
	
	\item The proposed SSCFL module aims to effectively learn and integrate bi-temporal differential feature representations. Specifically, it focuses on the change correlation between spatial-spectral features and the temporal dimension, learning the essential differential features of bi-temporal hyperspectral images to provide multiscale spatial-spectral differential representations for subsequent change detection, thus further enhancing the model's ability to capture subtle changes in hyperspectral images.
	
	\item An AFAF module is proposed to perform the final aggregation of key change features extracted from previous modules. This module aims to leverage the complementary information of multiple differential feature maps to generate highly discriminative spatial-spectral feature representations, thereby achieving accurate change detection.
\end{itemize}

The remainder of this paper is organized as follows. Section \ref{sec2} presents the details of the proposed CHDFFN architecture. Section \ref{sec3} describes the four experimental datasets, parameter settings, and experimental results, including ablation and comparative studies. Finally, Section \ref{sec4} concludes the paper and outlines directions for future research.

\section{Proposed method}
\label{sec2}
This section elaborates on the architecture of proposed cross-hierarchical difference feature fusion network, as illustrated in Fig. \ref{fig_1}. The framework primarily consists of three components: a multiscale feature extraction subnetwork, a spatial-spectral change feature learning module, and an advanced feature adaptive fusion module. The model progressively extracts multiscale spatial-spectral feature from the input bi-temporal HSI. Subsequently, the SSCFL module performs initial feature fusion and representation learning. Thereafter, the adaptive fusion module further refines and integrates these features, which are finally fed into an MLP classifier to perform change detection. The detailed design of each module will be introduced in the following subsections.
\begin{figure*}
	\centering
	\includegraphics[width=.9\textwidth]{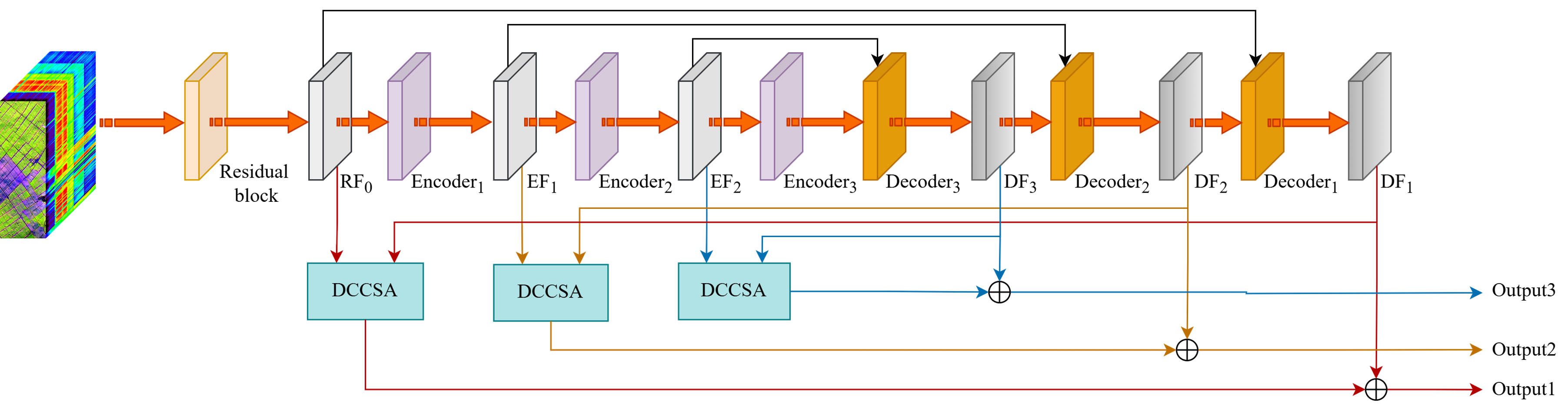}
	\caption{Structure of the multiscale feature extraction subnetwork.}
	\label{fig_2}
\end{figure*}
\begin{figure}
	\centering
	\includegraphics[width=.9\columnwidth]{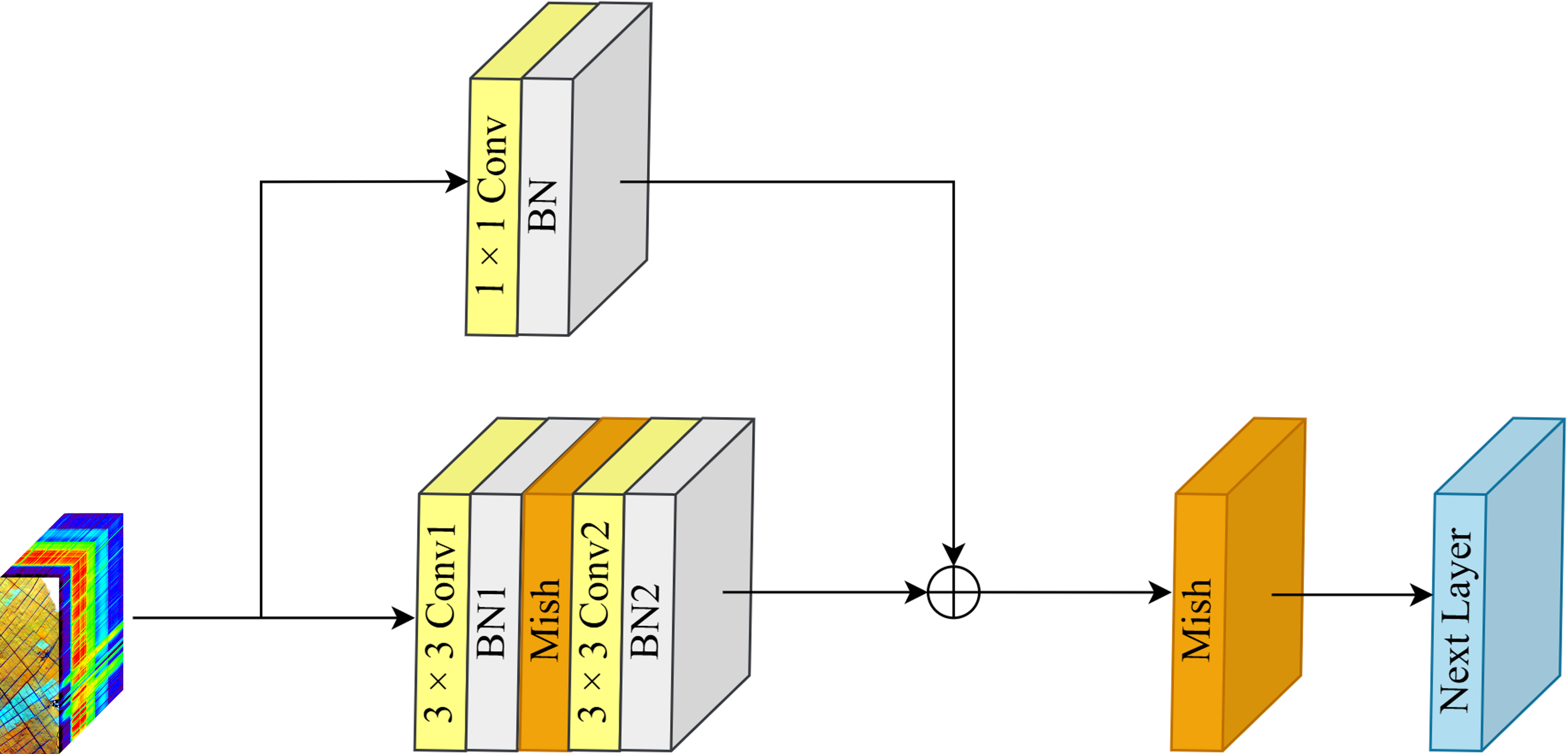}
	\caption{Structure of the residual block.}
	\label{fig_3}
\end{figure}

\subsection{Multiscale feature extraction subnetwork}
\label{subsec1}
\textit{1) Structure of the multiscale feature extraction subnetwork: }Due to their strong capabilities in data modeling and feature learning, CNNs are widely employed in HCD tasks to extract spatial-spectral\cite{ozdemir20233d} joint features. However, most existing methods integrate feature from only a single level, either at the input or output stage, which often results in the loss of critical information. Low-level features typically capture fine spatial details, such as textures and edges\cite{han2023hanet,yuan2022transformer}, while high-level features tend to encode abstract and coarse semantic information. This imbalance makes it difficult to generate accurate and well-defined boundaries. Therefore, the fusion of multiscale and multi-level information is essential for more precise representation of HSI feature.

In light of the above considerations, we propose a multiscale feature extraction subnetwork (MSFES) module. The structure of the module is illustrated in Fig. \ref{fig_2}, where RF, EF, and DF denote the output features from the residual block, encoder, and decoder layers, respectively. This subnetwork aims to extract and integrate multiscale hierarchical features from residual blocks, multiscale encoders, and decoders to enrich the spatial-spectral representations of hyperspectral data. The residual block performs initial preprocessing and encodes the spatial-spectral characteristics of the input. Subsequently, the multiscale encoder, consisting of convolutional layers with various kernel sizes, further encodes the extracted information to comprehensively capture spatial-spectral feature. The decoder then generates refined feature representations based on the joint features from the encoder. To improve the decoder’s ability to reconstruct detailed target information, a DCCSA module is embedded within the MSFES. This attention variant aims to extract and convey highly relevant joint feature information between low-level and high-level layers. Unlike conventional attention mechanisms, DCCSA incorporates enhanced dual-core channel attention and spatial attention components, offering more discriminative and informative attention weights.

\textit{2) Residual block: }As illustrated in Fig. \ref{fig_3}, the residual block consists of two consecutive 3×3 convolutional layers, each followed by batch normalization (BN) and the self-regularizing Mish activation function. A skip connection is incorporated in parallel, which adjusts the feature dimensions via a 1×1 convolution followed by BN. This design improves the diversity of feature extraction and enhances the training stability of the network. Given the input patch $h_I$, the final output of the residual block, denoted as $Out_r$, can be mathematically formulated as:
\begin{equation}  
	\begin{split}
		Out_r &= \sigma_{\textit{Mish}} \left( BN_2 \left( Conv_2 \left( \sigma_{\textit{Mish}} \left( BN_1 \left( Conv_1 (h_I) \right) \right) \right) \right) \right. \\
		&\quad + \left. BN \left( Conv (h_I) \right) \right)
	\end{split}
\end{equation}
Where $\sigma_{\textit{Mish}}(\cdot)$ denotes the Mish activation function, whose definition is as follows:
\begin{equation}
	Mish(x) = x \cdot \textit{tanh}(\textit{ln}(1 + e^x))
\end{equation}

\begin{figure*}
	\centering
	\includegraphics[width=.9\textwidth]{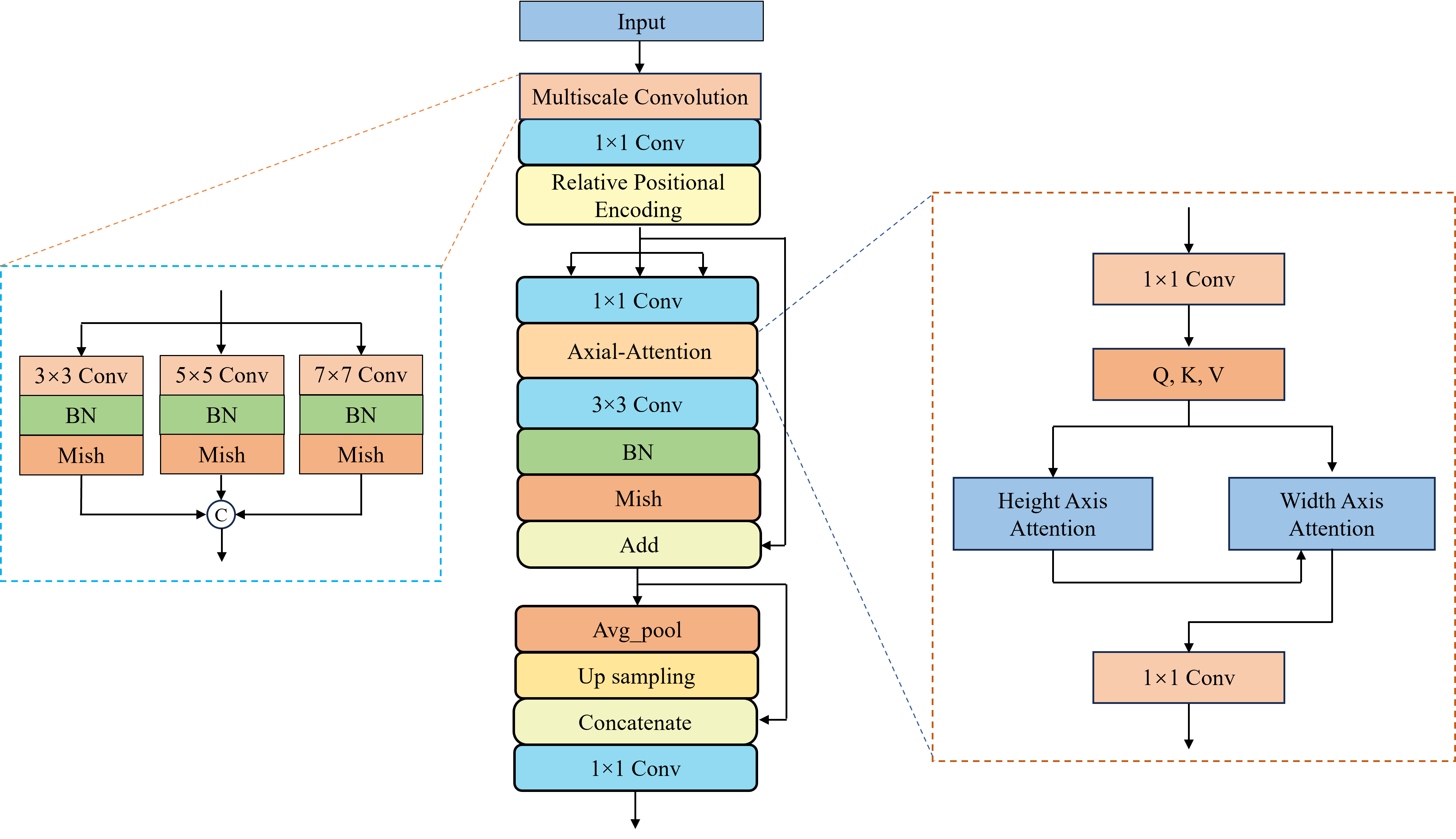}
	\caption{Structure diagram of the multiscale encoder. From left to right: multiscale convolution module, axial attention.}
	\label{fig_4}
\end{figure*}

\textit{3) Multiscale encoder module: }Fig. \ref{fig_4} presents the architecture of encoder module proposed in this paper. This module integrates multi-scale convolution, relative positional encoding, and axial attention mechanism, aiming to efficiently capture local and global contextual information. Specifically: the multi-scale convolution module captures multi-scale features through parallel convolutional layers with kernel sizes of 3×3, 5×5, and 7×7 respectively. Each convolutional layer is sequentially followed by a BN layer and a Mish activation function. Finally, the fusion of multi-scale features is achieved through a feature concatenation operation, so as to balance the capture of fine-grained and global information. For the input feature map \(x \in \mathbb{R}^{B \times C \times H \times W}\), the output of the k-th branch is:
\begin{equation}
	y_{\textit{k}} = \textit{Mish}(\textit{BN}(\textit{Conv}_{n \times n}(x)))
\end{equation}
Among them, \textit{B, C, H,} and \textit{w} denote the batch size, number of feature channels, feature map height, and feature map width respectively. represents the convolution operation with a kernel size of , and stands for BN. Here, \textit{n} = 3, 5, 7 and \textit{k} = 1, 2, 3. The final output of the multi-scale convolution module is:
\begin{equation}
	y = \textit{Concat}(y_1, y_2, y_3)
\end{equation}

The module introduces a relative positional encoding mechanism, which embeds positional information by modeling the relative distances between elements in the feature map, thereby enhancing the model's perception of spatial relationships. For any two positions (\textit{h}, \textit{w}) and (\textit{h'}, \textit{w'}), the calculations of the relative distance $\Delta \text{\textit{h}}$ in the height direction and the relative distance $\Delta \text{\textit{w}}$ in the width direction are shown in eq. \ref{eq.5} and \ref{eq.6}. For the convenience of calculating the encoding, the relative distances can be mapped to the non-negative integer range. Let the maximum relative distances in the height and width directions be H and W respectively (usually set to the maximum size values of the feature map), their mathematical expressions are shown in eq. \ref{eq.7} and \ref{eq.8}.
\begin{equation}
	\label{eq.5}
	\Delta \text{\textit{h}} = \text{\textit{h}} - \text{\textit{h}}' \\  
\end{equation}
\begin{equation}
	\label{eq.6}
	\Delta \text{\textit{w}} = \text{\textit{w}} - \text{\textit{w}}' \\  
\end{equation}
\begin{equation}
	\label{eq.7}
	\Delta \text{\textit{h}}' = \Delta \text{\textit{h}} + \text{\textit{H}}_{\text{max}} \\  
\end{equation}
\begin{equation}
	\label{eq.8}
	\Delta \text{\textit{w}}' = \Delta \text{\textit{w}} + \text{\textit{W}}_{\text{max}}  
\end{equation}

Therefore, for any relative position \((\Delta h, \Delta w)\) in the feature, the odd-even channel encoding vectors \(\textit{PosE}_h\) and \( \textit{PosE}_w\) in the height direction and width direction are calculated as shown in Eq. \ref{eq:9}, and the final output expression \( x_{\textit{pos}}\) is shown in Eq. \ref{eq:10}.
\begin{equation}
	\label{eq:9}  
	\textit{PosE}_{h,w}(\Delta \text{\textit{h}}, \Delta \text{\textit{w}}, c) =
	\begin{cases}
		\sin\!\left( \frac{\Delta \text{\textit{h}}}{10000^{2k/c}} \right), & c = 2k \\[6pt]
		\cos\!\left( \frac{\Delta \text{\textit{h}}}{10000^{2k/c}} \right), & c = 2k + 1 \\[6pt]
		\sin\!\left( \frac{\Delta \text{\textit{w}}}{10000^{2k/c}} \right), & c = \frac{C}{2} + 2k \\[6pt]
		\cos\!\left( \frac{\Delta \text{\textit{w}}}{10000^{2k/c}} \right), & c = \frac{C}{2} + 2k + 1
	\end{cases}
\end{equation}
Where \(k = 0, 1, 2, \dots, (C/4) - 1\).
\begin{equation}
	\label{eq:10}  
	x_{\text{\textit{pos}}} = x + \textit{Concat}\,(\textit{PosE}_{\text{\textit{h}}}(\Delta \text{\textit{h}}), \textit{PosE}_{\text{\textit{w}}}(\Delta \text{\textit{w}}))
\end{equation}

A 1×1 convolution is used to perform dimension transformation on the features fused with positional encoding, and then the transformed feature are input into the spatially axial attention module to generate query (Q), key (K), and value (V). After that, attention weights are calculated along the height axis and width axis respectively, and convolutional fusion is performed on the two attention features, so as to achieve the purpose of enhancing the position-aware capability of features and capturing long-range dependencies across spatial dimensions. For the height-axis attention, the calculation process of its output \(\textit{Output}_{h}^{h\textit{-}axis}\) is as follows:
\begin{equation}
	C_h = \frac{C}{H_d} \label{eq:11}
\end{equation}
\begin{equation}
	Q_h = Q[:, \text{\textit{h}} \cdot C_h: (\text{\textit{h}} + 1) \cdot C_h, :, :] \label{eq:12}
\end{equation}
\begin{equation}
	K_h = K[:, \text{\textit{h}} \cdot C_h: (\text{\textit{h}} + 1) \cdot C_h, :, :] \label{eq:13}
\end{equation}
\begin{equation}
	V_h = V[:, \text{\textit{h}} \cdot C_h: (\text{\textit{h}} + 1) \cdot C_h, :, :] \label{eq:14}
\end{equation}
\begin{equation}
	Q_h^{h\text{-}axis} = \begin{aligned}[t]
		\text{reshape}(&\text{permute}(Q_h, (0,1,3,2,4)), \\
		&(B \cdot H_d \cdot W, H, C_h))
	\end{aligned} \label{eq:15}
\end{equation}
\begin{equation}
	K_h^{h\text{-}axis} = \begin{aligned}[t]
		\text{reshape}(&\text{permute}(K_h, (0,1,3,2,4)), \\
		&(B \cdot H_d \cdot W, H, C_h))
	\end{aligned} \label{eq:16}
\end{equation}
\begin{equation}
	V_h^{h\text{-}axis} = \begin{aligned}[t]
		\text{reshape}(&\text{permute}(V_h, (0,1,3,2,4)), \\
		&(B \cdot H_d \cdot W, H, C_h))
	\end{aligned} \label{eq:17}
\end{equation}
\begin{equation}
	\textit{Attn}_h = \text{softmax}\left( \frac{Q_h^{h\text{-}axis} \cdot (K_h^{h\text{-}axis})^T}{\sqrt{C_h}} + b_h \right) \label{eq:18}
\end{equation}
\begin{equation}
	\textit{Output}_h^{h\text{-}axis} = \begin{aligned}[t]
		\text{permute}(&\text{reshape}(\text{Attn}_h \cdot V_h^{h\text{-}axis}, \\
		&(B \cdot H_d \cdot W, H, C_h)), (0,1,3,2,4))
	\end{aligned} \label{eq:19}
\end{equation}
Where \(H_d\) is the number of attention heads, \(C_h\) is the number of channels contained in a single attention head, \(Q_h, K_h, V_h\) \\are the Q, K, V values of each attention head, and \(Q_h^{h\text{-}axis}, \\K_h^{h\text{-}axis}, V_h^{h\text{-}axis}\) are the Q, K, V values in the height direction.

Similarly, the calculation process of the width-axis attention output \(\textit{Output}_{h}^{w\text{-}axis}\) can be described as follows:
\begin{equation}
	Q_{\text{\textit{h}}}^{w\text{-}axis} = \begin{aligned}[t]
		\text{reshape}(&\text{permute}(\text{Output}_{\text{\textit{h}}}^{h\text{-}axis}, (0,1,2,3,4)), \\
		&(B \cdot H_d \cdot H, W, C_{\text{\textit{h}}}) )
	\end{aligned} \label{eq:20}
\end{equation}
\begin{equation}
	K_{\text{\textit{h}}}^{w\text{-}axis} = \begin{aligned}[t]
		\text{reshape}(&\text{permute}(K_{\text{\textit{h}}}, (0,1,2,3,4)), \\
		&(B \cdot H_d \cdot H, W, C_{\text{\textit{h}}}) )
	\end{aligned} \label{eq:21}
\end{equation}
\begin{equation}
	V_{\text{\textit{h}}}^{w\text{-}axis} = \begin{aligned}[t]
		\text{reshape}(&\text{permute}(V_{\text{\textit{h}}}, (0,1,2,3,4)), \\
		&(B \cdot H_d \cdot H, W, C_{\text{\textit{h}}}) )
	\end{aligned} \label{eq:22}
\end{equation}
\begin{equation}
	\textit{Attn}_{\text{\textit{w}}} = \text{softmax}\left( \frac{Q_{\text{\textit{h}}}^{w\text{-}axis} \cdot (K_{\text{\textit{h}}}^{w\text{-}axis})^T}{\sqrt{C_{\text{\textit{h}}}}} + b_{\text{\textit{w}}} \right) \label{eq:23}
\end{equation}
\begin{equation}
	\textit{Output}_{\text{\textit{h}}}^{w\text{-}axis} = \begin{aligned}[t]
		\text{permute}(&\text{reshape}(\text{Attn}_{\text{\textit{w}}} \cdot V_{\text{\textit{h}}}^{w\text{-}axis}, \\
		&(B \cdot H_d \cdot H, W, C_{\text{\textit{h}}}) ), (0,1,2,3,4))
	\end{aligned} \label{eq:24}
\end{equation}
Where \(Q_h^{w\text{-}axis}, K_h^{w\text{-}axis}, V_h^{w\text{-}axis}\) are the Q, K, V values in the width direction.

Finally, the output of the spatially separable axial attention is:
\begin{equation}
	\text{\textit{Attn}}_{\text{\textit{ssa}}} = \text{Conv}_{1 \times 1}\!\bigl(\text{\textit{output}}_{\text{\textit{h}}}^{w\text{-}axis}\bigr)
	\label{eq:attn_ssa}  
\end{equation}
The fused features sequentially pass through a 3×3 convolution, a BN layer, and a Mish activation function to output spatially axial attention features. This feature and the position-encoded feature establish a residual connection through the ``\textit{Add} '' operation to achieve feature reuse. Finally, through a series of processes including average pooling, upsampling, feature concatenation, and the final 1×1 convolution, the process of feature extraction and fusion within the encoder is completed.

\begin{figure*}
	\centering
	\includegraphics[width=.9\textwidth]{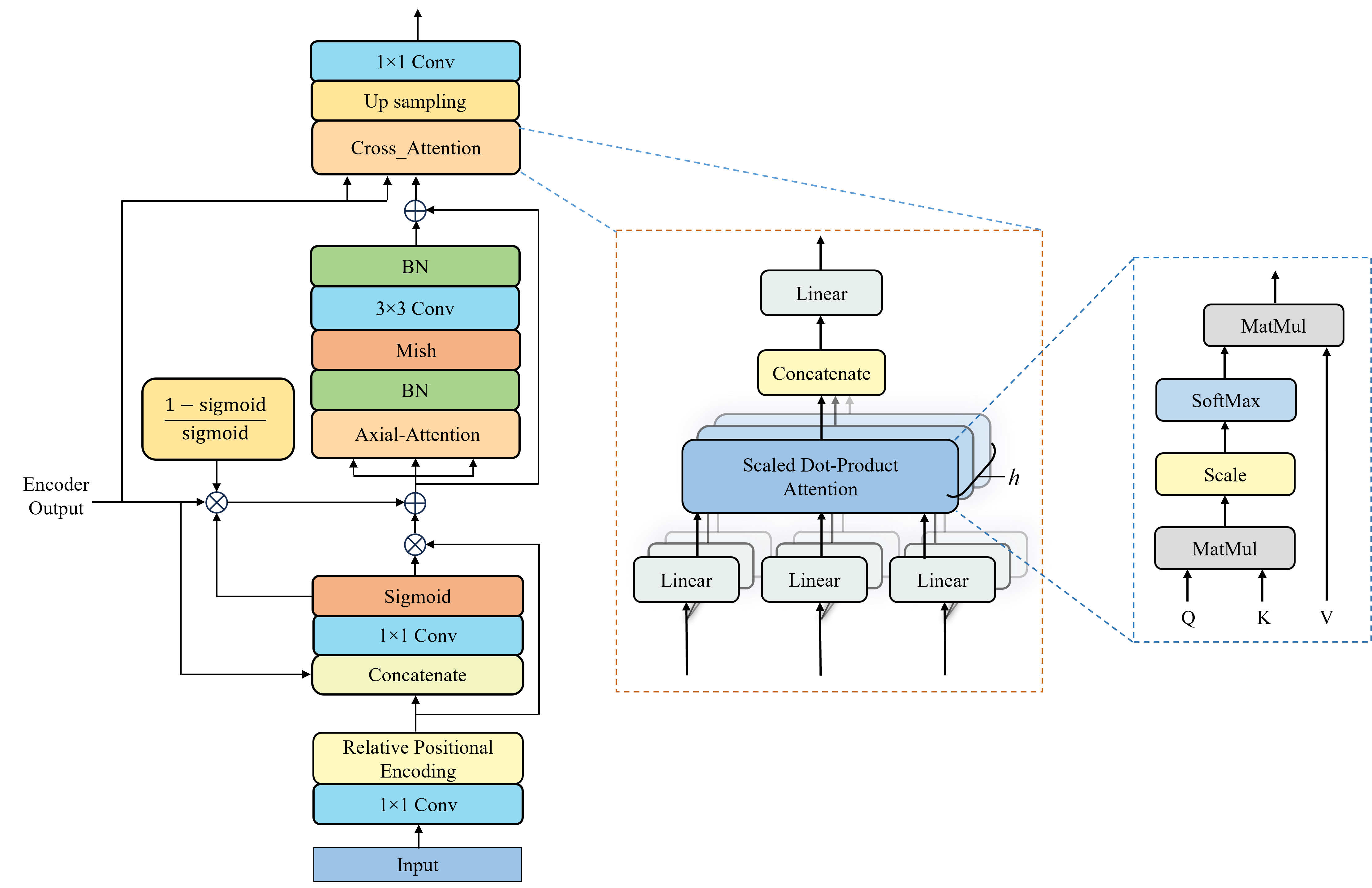}
	\caption{Structure of the decoder. From left to right: decoder, multi-head self-attention, and scaled dot-product attention.}
	\label{fig_5}
\end{figure*}

\textit{4) Decoder module: }Fig. \ref{fig_5} presents the structural diagram of the decoder module proposed in this paper. This module incorporates cross-attention, depth-wise separable axial attention, and relative positional encoding, aiming to achieve multi-dimensional feature enhancement and efficient fusion of cross-source information.

The input features are first passed through a 1×1 convolution to adjust the channel dimension. Subsequently, positional information is injected via relative positional encoding and concatenated with the encoder output. After convolution operation and Sigmoid activation, attention weights are generated, and the encoder output is weighted through element-wise multiplication to realize gated interaction of cross-level features. Assuming the input feature is \(X_{in} \in \mathbb{R}^{C \times H \times W}\) and the encoder output feature is \(X_{enc} \in \mathbb{R}^{C_{enc} \times H_{enc} \times W_{enc}}\), then:
\begin{equation}
	\text{\textit{X}}_1 = \textit{Conv}_{1 \times 1}^{\text{\textit{init}}}(\text{\textit{X}}_{\text{\textit{in}}}) + \textit{RPE}(\text{\textit{X}}_{\text{\textit{in}}}) \label{eq:x1}
\end{equation}
\begin{equation}
	\alpha = \textit{Sigmoid}(\textit{Conv}_{1 \times 1}^{\text{\textit{gate}}}(\textit{Concat}(\text{\textit{X}}_1, \text{\textit{X}}_{\text{\textit{enc}}}))) \label{eq:alpha}
\end{equation}
\begin{equation}
	\text{\textit{X}}_{\text{\textit{enc}-\textit{weighted}}} = \text{\textit{X}}_{\text{\textit{enc}}} \otimes (1 - \alpha) \label{eq:x_enc_weighted}
\end{equation}
Where \(Conv_{1 \times 1}^{init}\) and \(Conv_{1 \times 1}^{gate}\) denote the 1×1 convolution operations in the initial and gating branches respectively, \(\textit{RPE}(\cdot)\) is the relative positional encoding function, and \(\otimes\) represents element-wise multiplication.

The weighted and residual branch features are fused through addition, and then fed into the spatially separable axial attention module to capture dependencies in the spatial dimension. Subsequently, they sequentially pass through BN, Mish activation, 3×3 convolution, and another BN layer, and complete the residual connection through addition again, thereby enhancing the non-linear expression of features and the efficiency of gradient transmission.
\begin{equation}
	\text{\textit{X}}_2 = \text{\textit{X}}_1 \otimes \alpha + \text{\textit{X}}_{\text{\textit{enc}-\textit{weighted}}} \label{eq:x2}
\end{equation}
\begin{equation}
	\text{\textit{X}}_3 = \textit{Axial}_{\text{\textit{attention}}}(\text{\textit{X}}_2) \label{eq:x3}
\end{equation}
\begin{equation}
	\text{\textit{X}}_4 = \textit{BN}(\textit{Conv}_{3 \times 3}(\textit{Mish}(\textit{BN}(\text{\textit{X}}_3)))) \label{eq:x4}
\end{equation}
\begin{equation}
	\text{\textit{X}}_5 = \text{\textit{X}}_2 + \text{\textit{X}}_4 \label{eq:x5}
\end{equation}
Among them, \(X_2, X_3, X_4, X_5\) are the gated interaction output feature, the spatially separable axis attention output feature, the locally enhanced feature of \(X_3\), and the cross-layer fusion feature, respectively.

Subsequently, the features enter the cross-attention submodule: this submodule adopts the multi-head scaled dot-product attention mechanism. Each head generates Q, K, and V respectively through the Linear layer. Q and K perform matrix multiplication, and after scaling, the attention weights are obtained through SoftMax. Then, matrix multiplication is performed with V. Finally, the outputs of all attention heads are concatenated and integrated through the Linear layer before being output.
\begin{equation}
	\text{\textit{d}}_k = \frac{\text{\textit{C}}_5}{\text{\textit{h}}} \label{eq:d_k}
\end{equation}
\begin{equation}
	X_6 = \begin{aligned}[t]
		L_{\text{out}} ( \text{Concat}_{i=1}^h ( ( \text{softmax} \, \bigl( \frac{L_Q^i(X_5) \cdot (L_K^i(X_5))^T}{\sqrt{d_k}} \bigr) \\
		\cdot L_V^i(X_5) ) ) )
	\end{aligned}
	\label{eq:x6}
\end{equation}
Among them, \(h, d_k, C_5\) are the number of attention heads, the dimension size of each attention head, and the number of channels of \(X_5\), respectively, and \(L^i_{Q,K,V}\) represents the Q, K, V values generated by linear transformation.

Features output by cross-attention restore spatial resolution through upsampling, and finally adjust the channel dimension via 1×1 convolution to complete feature enhancement and transmission of the entire module. The final output \(X_{out}\) is described as follows:
\begin{equation}
	\text{\textit{X}}_{\text{\textit{out}}} = \text{Conv}_{1 \times 1}(\text{Up}(\text{\textit{X}}_6))
	\label{eq:x_out}  
\end{equation}
Where \(Up(\cdot)\) denotes the upsampling function.
\begin{figure*}
	\centering
	\includegraphics[width=.9\textwidth]{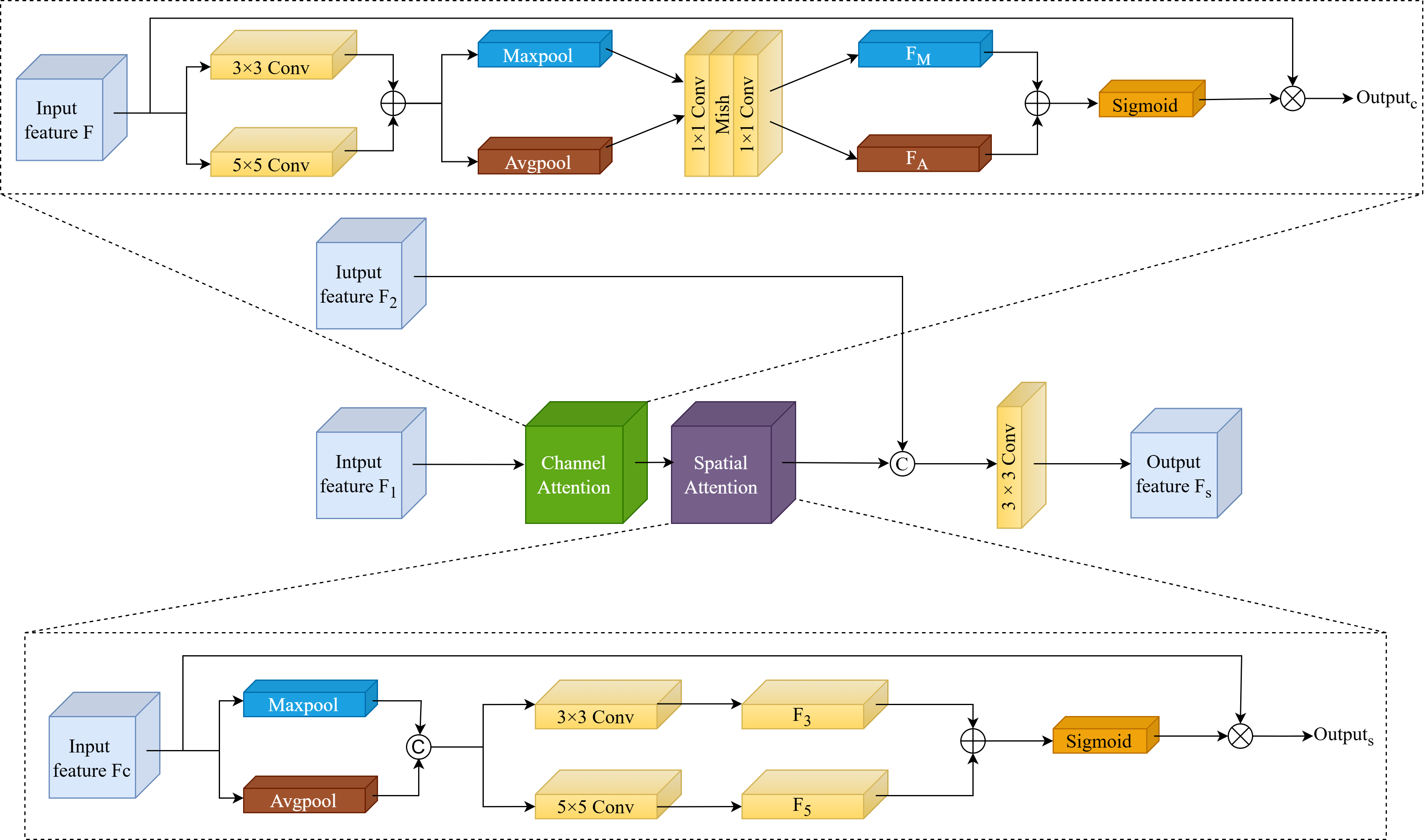}
	\caption{Structure of the DCCSA module.}
	\label{fig_6}
\end{figure*}

\textit{5) Dual-core channel-spatial attention module: }To enhance attention weights for better highlighting of important features, we introduce two parallel convolutional layers into both the channel and spatial attention branches, forming a novel dual-core attention module. The module employs a cascaded design combining enhanced channel and spatial attention mechanisms, enabling joint processing of input features. Its structure is shown in Fig. \ref{fig_6}. The input feature tensor $F$ first passes through the channel attention branch, where it undergoes parallel 3×3 and 5×5 convolutions to extract and fuse multiscale features, followed by a shared MLP layer for learning discriminative representations. Finally, the MLP outputs are combined via element-wise addition and passed through a Sigmoid activation function to generate the attention weight $W$. The input $F$ is then multiplied by $W$ to obtain the channel attention output $Output_{\textit{c}}$.
\begin{equation}
	\textit{Output}_{\textit{c}} = F_1 \cdot \sigma_{\textit{Sigmoid}}(W)
\end{equation}
Where $F$ is the input feature, $W$ is the attention weight matrix, and $\sigma_{\textit{Sigmoid}}(\cdot)$ denotes the activation operation performed using the Sigmoid activation function.
\begin{align}
	\begin{split}
		W &= \textit{Conv}_{1 \times 1}\left( \sigma_{\textit{Mish}}\left( \textit{Conv}_{1 \times 1}\left( \textit{MaxPool}(F_{conv}) \right) \right) \right) + \\
		&\quad \textit{Conv}_{1 \times 1}\left( \sigma_{\textit{Mish}}\left( \textit{Conv}_{1 \times 1}\left( \textit{AvgPool}(F_{conv}) \right) \right) \right)
	\end{split}
\end{align}
\begin{equation}
	F_{\textit{conv}} = \textit{Conv}_{3 \times 3}(F) + \textit{Conv}_{5 \times 5}(F)
\end{equation}

The spatial attention module takes the channel-refined feature $F_{\textit{c}}$ as input. To aggregate spatial information, both max pooling and average pooling operations are applied, and their outputs are concatenated to form a multiscale feature representation. These concatenated features are then processed by 3×3 and 5×5 convolutional layers, respectively, to capture details from different receptive fields. After feature fusion, spatial attention weights are generated via the Sigmoid activation function and used to weight $F_{\textit{c}}$, resulting in the spatial attention output $F_{\textit{s}}$. By leveraging a decoupled structure of channel and spatial attention mechanisms, this module enables adaptive enhancement of both channel-wise and spatial feature representations, thereby improving the model’s ability to capture critical information. The output is computed as follows:
\begin{align}
	\begin{split}
		F_s &= \textit{Conv}_{3 \times 3} \left( \textit{Concat} \left( F_2,\ F_c \cdot \sigma_{\textit{Sigmoid}} \left( \right. \right. \right. \\
		&\quad \left. \left. \left. \textit{Conv}_{3 \times 3}(F_{pool}) + \textit{Conv}_{5 \times 5}(F_{pool}) \right) \right) \right)
	\end{split} \\
	F_{pool} &= \textit{Concat} \left( \textit{MaxPool}(F_c),\ \textit{AvgPool}(F_c) \right)
\end{align}
Where $F_c$ is the output value $Output_c$ of the channel attention module, $Concat(\cdot)$ represents the concatenation operation, and $\sigma_{\textit{Sigmoid}}(\cdot)$ denotes the activation operation of the activation function.

\subsection{Spatial-Spectral Change Feature Learning Module}
\label{subsec2}
\begin{figure}
	\centering
	\includegraphics[width=.95\columnwidth]{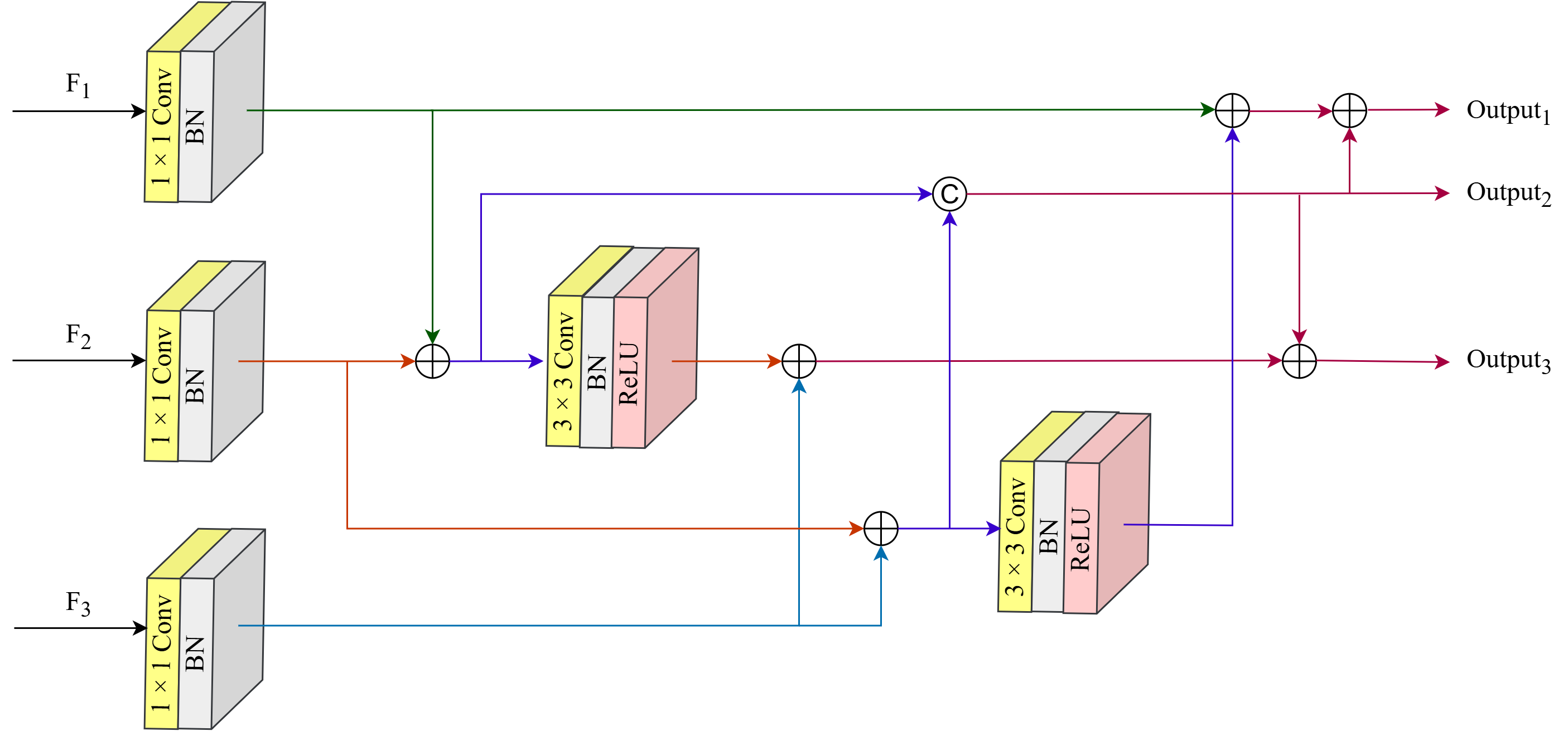}
	\caption{Structure of the SSCFL module.}
	\label{fig_7}
\end{figure}

In bi-temporal HSI processing, the input image patches are first passed through a multiscale feature extraction subnetwork, resulting in three feature maps with varying levels and multiple scales. These feature maps consist of both change and invariant components. Among them, the change components possess stronger representational capacity for fine-grained changes and directly contribute to the core objective of change pixel detection. However, existing studies often overlook the targeted learning of change components within spatial-spectral joint features during change detection. To address this issue, this paper proposes a SSCFL module, which effectively captures subtle change features through a multi-dimensional feature interaction mechanism. The detailed architecture of the proposed module is illustrated in Fig. \ref{fig_7}.

The module adopts a multi-branch parallel architecture that integrates 1×1 convolutions, multi-stage 3×3 convolutions, and a dense residual connection mechanism to construct a spatial-spectral-temporal joint feature learning framework. Specifically, each branch generates multi-granularity representations by extracting features at different scales. Then, deeper features are obtained through stage-wise convolutions, and feature maps from different levels are fused using cross-layer residual connections. The three output expressions of the module are defined as follows:
\begin{equation}
	\begin{split}
		Output_1 &= \left( BN(Conv_{1 \times 1}(F_1)) \right. \\
		&\quad \left. + \sigma_{\textit{ReLU}}(BN(Conv_{3 \times 3}(X_2))) \right) \\
		&\quad + Concat(X_1, X_2)
	\end{split}
	\label{eq:output1}
\end{equation}
\begin{equation}
	\textit{Output}_2 = Concat(X_1, X_2)
\end{equation}
\begin{equation}
	\begin{split}
		\textit{Output}_3 &= \sigma_{\textit{ReLU}}(BN(Conv_{3 \times 3}(X_1))) + BN(Conv_{1 \times 1}(F_3)) \\
		&\quad + Concat(X_1, X_2)
	\end{split}
\end{equation}
\begin{equation}
	X_1 = BN(Conv_{1 \times 1}(F_1)) + BN(Conv_{1 \times 1}(F_2))
\end{equation}
\begin{equation}
	X_2 = BN(Conv_{1 \times 1}(F_2)) + BN(Conv_{1 \times 1}(F_3))
\end{equation}
where $F_1$, $F_2$, and $F_3$ denote the three inputs of the module(feature output value of the feature extraction subnetwork), $BN(\cdot)$ stands for the batch normalization operation.

This design strategy offers three advantages: First, the multi-branch parallel structure effectively captures initial features at different scales, enhancing information diversity through multi-path feature extraction. Second, the dense residual connection mechanism constructs shortcut paths for gradient propagation, significantly alleviating the gradient vanishing problem in deep networks and supporting the training of deeper models. Third, the cross-layer feature fusion strategy realizes complementary reuse of features at different levels, preserving shallow detail information while fusing deep semantic features, thereby enhancing the robustness of feature expression and improving the model's ability to model complex change patterns.

\subsection{Advanced Feature Adaptive Fusion Module}
\label{subsec3}
\begin{figure}
	\centering
	\includegraphics[width=.95\columnwidth]{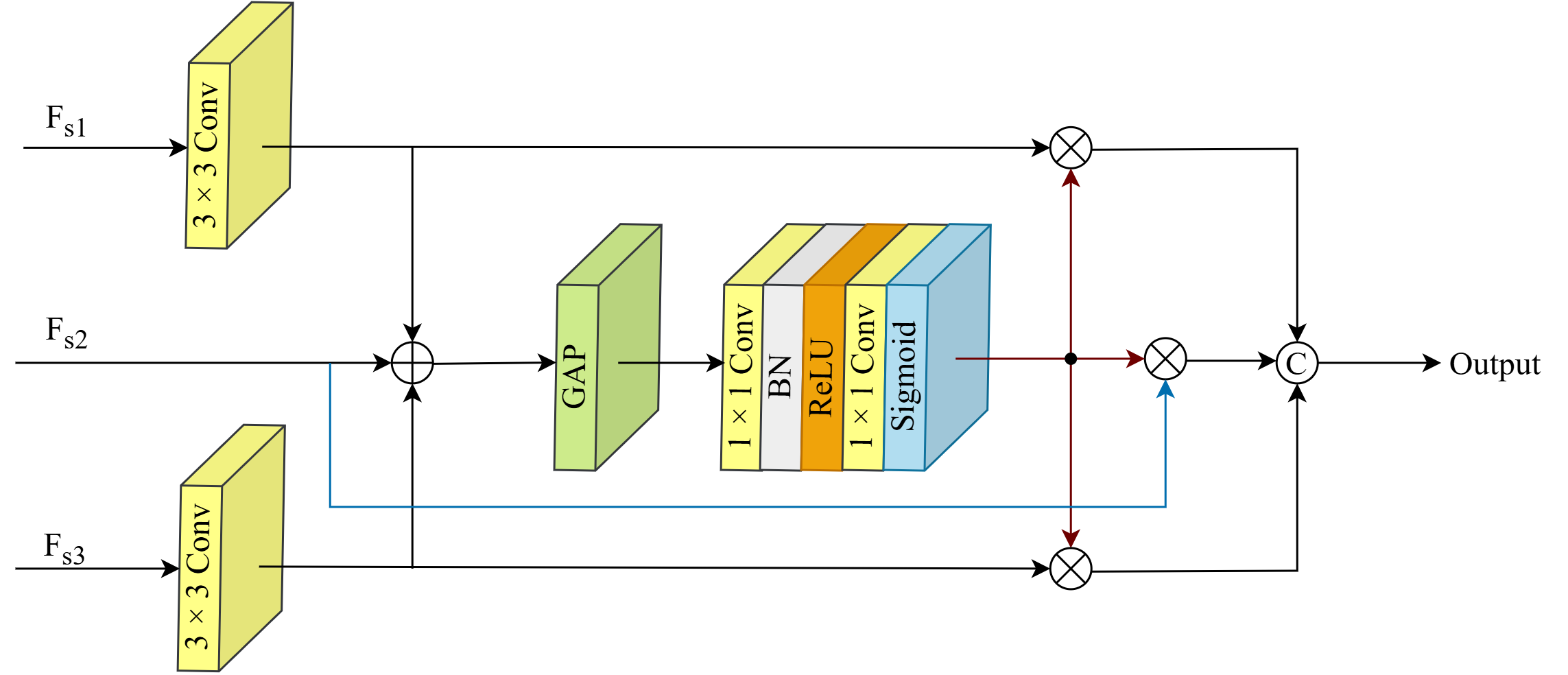}
	\caption{Structure of the AFAF module.}
	\label{fig_8}
\end{figure}

After being processed by the SSCFL module, three feature maps containing information at different spatial scales are generated. In traditional approaches, these multiscale features are typically fed directly into a fully connected layer for fusion, with final change detection performed using activation functions such as Sigmoid or SoftMax. However, this straightforward integration often results in suboptimal detection performance due to insufficient emphasis on salient information. As previously discussed, attention mechanisms help focus on important features and have been increasingly applied\cite{wang2021ssa} in the field of HCD. For example, Shangguan Y et al.\cite{shangguan2024multiscale} proposed an ACSP attention structure to explore the multiscale contextual information of coarse change features. ACSP employs contextual self-attention (CSA) branches with varying dilation rates to extract multiscale semantics, thereby improving the model’s ability to adaptively perceive and capture relevant changes. To address these limitations, this study introduces an adaptive feature aggregation mechanism that complements features across different scales and reinforces key information through dynamic weight allocation. The architectural diagram is shown in Fig. \ref{fig_8}.

The AFAF module achieves adaptive fusion of multiscale features by incorporating an attention mechanism. First, 3×3 convolutions are applied to the input features $F_{s1}$ and $F_{s3}$ to enhance local feature representation. Then, the convolution output of $F_{s1}$ and the original $F_{s2}$ features are compressed into channel-level statistics using Global Average Pooling (GAP). These statistics are subsequently passed through two sequential 1×1 convolution layers (with Batch Normalization and ReLU activation in between) for dimensionality reduction and transformation. After that, normalized attention weights are generated via the Sigmoid function. Each weight is applied to the corresponding feature—$F_{s1}$, $F_{s2}$, and $F_{s3}$—via channel-wise multiplication, enabling dynamic enhancement of important features. Finally, the three weighted features are concatenated along the channel dimension to produce the fused feature map. By adaptively adjusting the contribution of each input, this module improves the effectiveness of multiscale feature fusion and enhances overall feature representation.
\begin{align}
	\begin{split}
		\textit{Output} &= \textit{Concat}\left( \left( \textit{Conv}_{3 \times 3}(F_{s1}) \times W \right), \right. \\
		&\quad \left. \left( F_{s2} \times W \right), \left( \textit{Conv}_{3 \times 3}(F_{s3}) \times W \right) \right)
	\end{split} \\
	\begin{split}
		W &= \sigma_{\textit{Sigmoid}}\left( \textit{Conv}_{1 \times 1}\left( \sigma_{\textit{ReLU}}\left( BN\left( \right. \right. \right. \right. \\
		&\quad \left. \left. \left. \left. \textit{Conv}_{1 \times 1}\left( GAP(F) \right) \right) \right) \right) \right)
	\end{split} \\
	F &= \textit{Conv}_{3 \times 3}(F_{s1}) + F_{s2} + \textit{Conv}_{3 \times 3}(F_{s3})
\end{align}
where $W$ is the attention weight matrix,$F_{s1}$,$F_{s2}$,$F_{s3}$ are the output values of the SSCFL module respectively, and $F$ is the fused feature.

After multi-stage feature extraction by the preceding module, a feature map capturing multiscale change information is generated. This feature map is then passed through a fully connected layer to produce a probability distribution vector, which is subsequently processed by a classifier to obtain the final category label of the input data block, thereby achieving accurate change detection. The final result of HCD can be expressed as follows:
\begin{equation}
	y_n = \sigma_{\textit{softmax}} \left( F_c \left( F_c \left( \textit{Conv}_{5 \times 5}(\textit{Output}) \right) \right) \right)
\end{equation}
where $y_n$ is the probability that the input sample is predicted to belong to class $n$, $F_c$ denotes the fully connected layer, and \textit{$Output$} is the output of the AFAF module.

\subsection{Loss Function}
\label{subsec4}
In the training of DL model, adopting an appropriate loss function is essential for optimizing the network architecture and guiding the feature extraction module to learn more discriminative representations. In CD tasks, the cross-entropy loss function measures the divergence between the predicted results and ground-truth labels, offering a clear gradient direction for parameter updates. In binary classification scenarios, its optimization is equivalent to maximizing the log-likelihood of the correct class. This property gives it better convergence performance than traditional loss functions such as mean squared error, particularly in pixel-wise classification tasks. Moreover, combining cross-entropy loss with the softmax activation function enables the model to produce outputs that align with probability distributions. This facilitates subsequent threshold segmentation or fine-grained classification, ultimately enhancing the accuracy and spatial coherence of the change detection results. It can be formally expressed as:
\begin{equation}
	Loss = -\frac{1}{N} \sum_{i=1}^{N} \left( y_i \log(y_n) + (1 - y_i) \log(1 - y_n) \right)
\end{equation}
where $N$ is the number of batch samples, $y_i$ is the true label pixel value, and $y_n$ is the predicted probability value belonging to class $n$.

\section{Experiment}
\label{sec3}
This section begins by briefly introducing the hyperspectral image change detection dataset and the evaluation metrics employed for quantitatively assessing the performance of the proposed CHDFFN method. Subsequently, the experimental results of the proposed algorithm are discussed in detail. The effectiveness of key components, including the multiscale feature extraction module and the adaptive feature fusion module, is verified through modular ablation experiments. In addition, several state-of-the-art HCD algorithms are selected for performance comparison and analysis, with the corresponding experimental hyperparameter settings also provided. Finally, the impacts of training sample proportion, batch size, and input image patch size on network performance are analyzed through controlled experiments.

\subsection{Datasets and Evaluation Metrics}
\label{subsec5}
\begin{figure}
	\centering
	\includegraphics[width=.9\columnwidth]{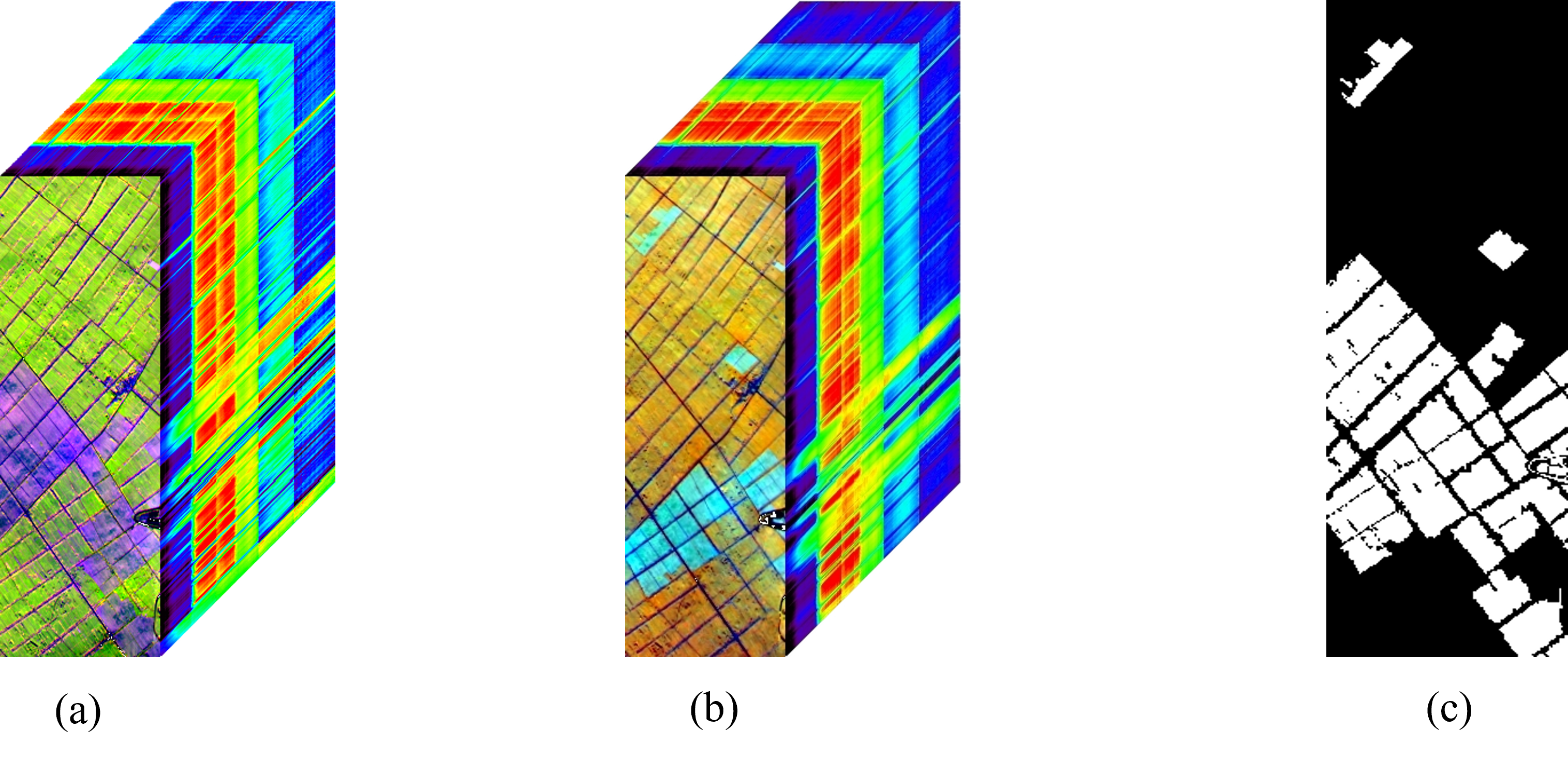}
	\caption{Farmland dataset. (a) Image acquired on May 3, 2006. (b) Image acquired on April 23, 2007. (c) Change ground truth.}
	\label{fig_9}
\end{figure}

\textit{1) Datasets: }The first dataset is the Yancheng dataset from Jiangsu Province, collected by the Hyperion sensor, hereafter referred to as `` Farmland ''. As shown in Fig. \ref{fig_9} (a) and \ref{fig_9} (b), two hyperspectral images, named T1 and T2, were acquired on May 3, 2006, and April 23, 2007, respectively. Each image has dimensions of 420 × 140 × 154, covering 154 spectral bands. Fig. \ref{fig_9} (c) presents the manually annotated ground truth of farmland changes, consisting of 40,417 unchanged pixels (black areas) and 18,383 changed pixels (white areas), reflecting the agricultural land transformations near Yancheng City, Jiangsu Province, China.

\begin{figure}
	\centering
	\includegraphics[width=.9\columnwidth]{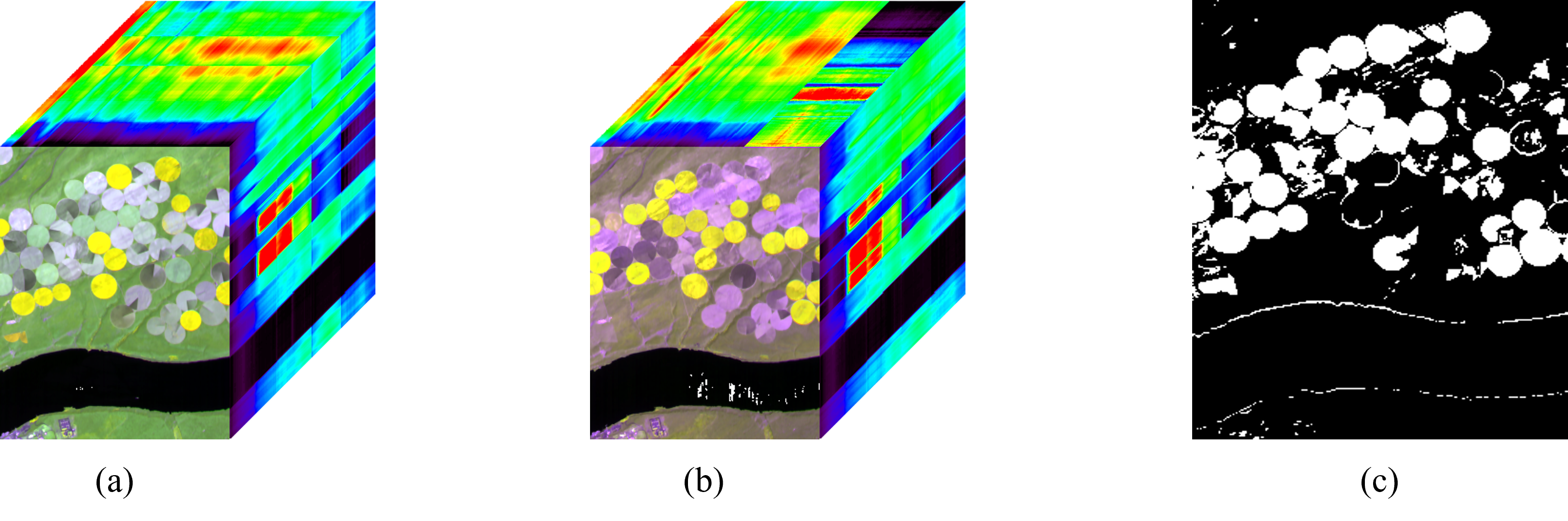}
	\caption{Hermiston2 dataset. (a) Image acquired on May 1, 2004. (b) Image acquired on May 8, 2007. (c) Change ground truth.}
	\label{fig_10}
\end{figure}

The second dataset, referred to as `` Hermiston2 '', was collected by the Hyperion sensor in irrigated farmlands located in Hermiston, Umatilla County, Oregon, USA. As shown in Fig. \ref{fig_10} (a) and \ref{fig_10} (b), the dataset has a spatial resolution of 307 × 241 pixels and comprises 154 spectral bands. T1 and T2 were acquired on May 1, 2004, and May 8, 2007, respectively. Fig. \ref{fig_10} (c) shows the ground truth of farmland changes, comprising 16,676 changed pixels (white areas) and 57,311 unchanged pixels (black areas), indicating the variations in irrigated farmlands over time in the Hermiston area.

\begin{figure}
	\centering
	\includegraphics[width=.9\columnwidth]{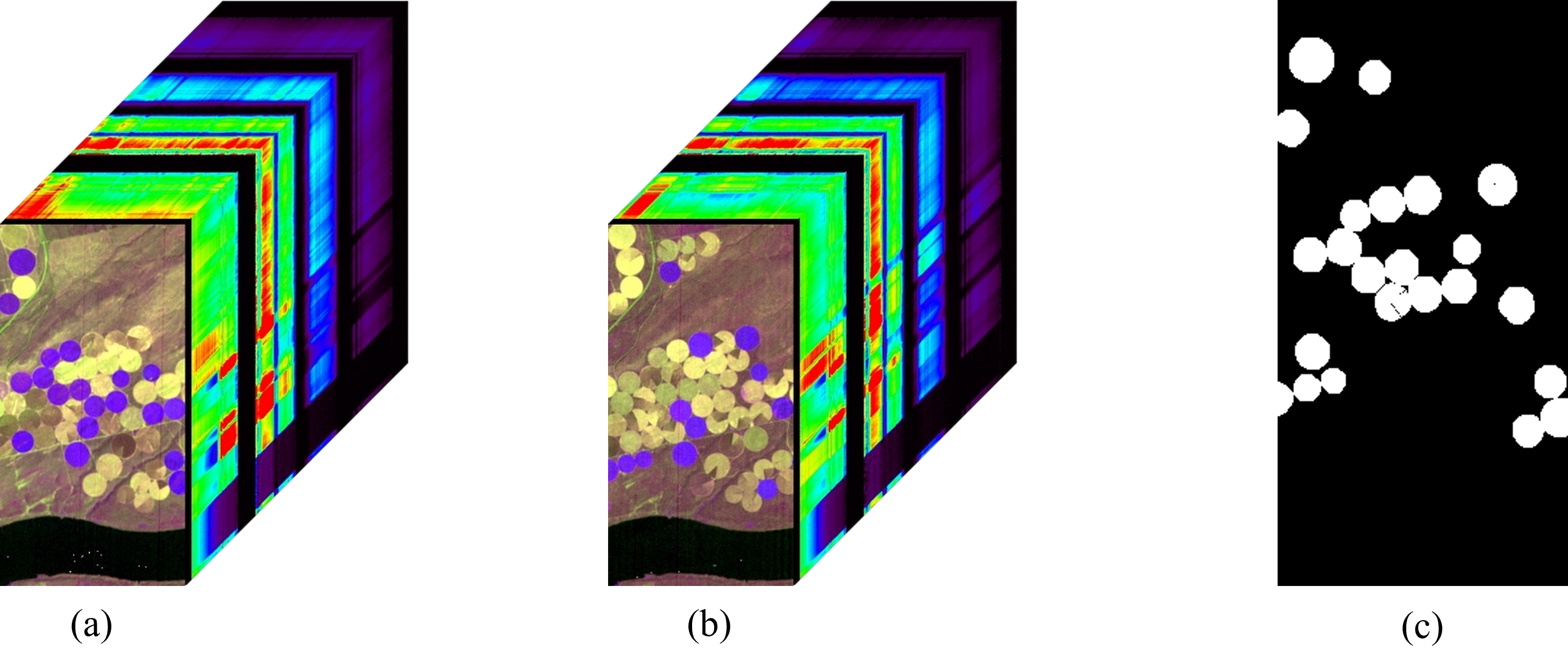}
	\caption{Hermiston dataset. (a) Image acquired on May 1, 2004. (b) Image acquired on May 8, 2007. (c) Change ground truth.}
	\label{fig_11}
\end{figure}

The third dataset, referred to as `` Hermiston '', focuses on circular farmlands. It covers agricultural areas in Hermiston, Umatilla County, Oregon, USA, with data collected by the Hyperion sensor on May 1, 2004, and May 8, 2007. As shown in Fig. \ref{fig_11} (a) and \ref{fig_11} (b), the dataset comprises 242 spectral bands with spatial dimensions of 390 × 200 pixels. Fig. \ref{fig_11} (c) presents the ground truth labels, indicating 9,986 changed pixels (white) and 68,014 unchanged pixels (black), which reflect actual changes in the circular farmland area.

\begin{figure}
	\centering
	\includegraphics[width=.9\columnwidth]{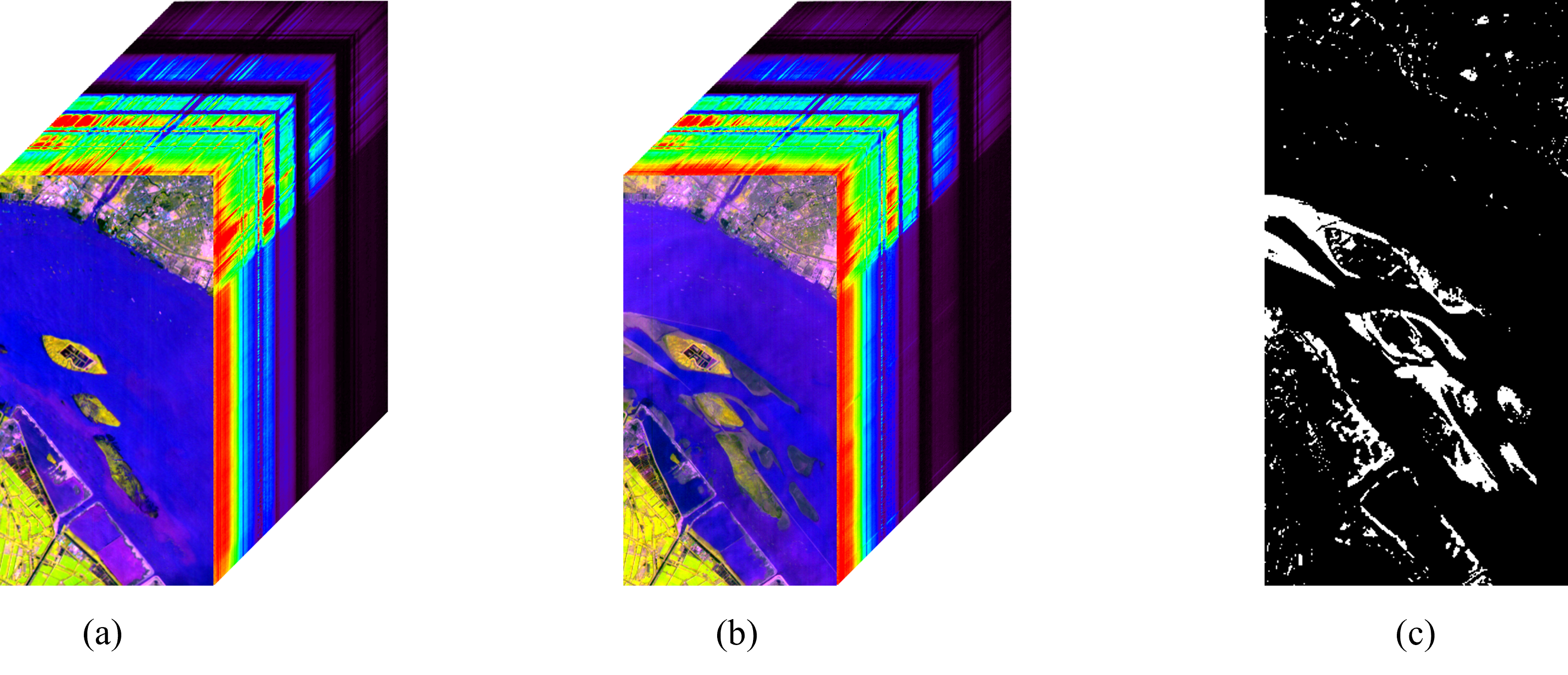}
	\caption{River dataset. (a) Image acquired on May 3, 2013. (b) Image acquired on December 31, 2013. (c) Change ground truth.}
	\label{fig_12}
\end{figure}

The fourth dataset, referred to as `` River '', represents a riverine region in Jiangsu Province, China. It comprises hyperspectral images acquired by the Hyperion sensor on May 3 and December 31, 2013. As illustrated in Fig. \ref{fig_12} (a) and \ref{fig_12} (b), this dataset includes 198 spectral bands with a spatial resolution of 463 × 241 pixels. Fig. \ref{fig_12} (c) presents the ground truth map of changes within the river area, comprising 101,885 unchanged pixels (black) and 9,698 changed pixels (white). The primary types of land cover change in this dataset involve variations in river sediment levels and channel morphology.

\textit{2) Evaluation Metrics: }To effectively assess the performance of the proposed method, we compare the predicted detection results with the ground-truth reference data. In this study, Overall Accuracy (OA) and the Kappa Coefficient (KC) are adopted as the primary evaluation metrics, while Precision (Pr), Recall (Re), and F1-score (F1) are employed as supplementary metrics. These metrics\cite{tang2021unsupervised,qin2024domain,ping2025Hypergraph,deng2024gated} jointly provide a comprehensive evaluation of the proposed algorithm.

OA represents the proportion of correctly predicted samples by the classification model to the total number of samples, reflecting the overall classification accuracy of the model. A larger value indicates a better overall classification effect.
\begin{equation}
	OA = \frac{TP + TN}{TP + FP + TN + FN}
\end{equation}

KC is used to measure the consistency between the prediction results of a classification model and the true results, taking into account the impact of chance factors in the classification results. Compared with Overall Accuracy, it can evaluate model performance more accurately. A larger value indicates a stronger consistency between the model's classification results and the true labels, and this consistency excludes the interference of random factors.
\begin{equation}
	KC = \frac{OA - p_e}{1 - p_e}
\end{equation}
Where the calculation process of the expected accuracy $p_e$ is as follows:
\begin{equation}
	p_e = \frac{(TP \times FN) + (TP \times FP) + (TN \times FN) + (TN \times FP)}{N^2}
\end{equation}

Pr refers to the proportion of true positive samples among those predicted as positive by the model, measuring the accuracy of the model in predicting positive classes. A larger value indicates a higher proportion of truly positive samples among those predicted as ``positive'' by the model, i.e., ``fewer false positives''.
\begin{equation}
	Pr = \frac{TP}{TP + FP}
\end{equation}

Re, also known as sensitivity or true positive rate, refers to the proportion of samples correctly predicted as positive among all actually positive samples, reflecting the model's ability to identify positive samples. A larger value indicates a higher proportion of truly ``positive'' samples that are successfully identified by the model, i.e., ``fewer false negatives''.
\begin{equation}
	Re = \frac{TP}{TP + FN}
\end{equation}

The F1 is the harmonic mean of precision and recall. It comprehensively considers the model's precision and recall, enabling a more holistic evaluation of model performance. A higher F1-score indicates a better balance between the model's Pr and Re, as well as a superior overall performance of the two.
\begin{equation}
	F1 = \frac{2PR}{P + R}
\end{equation}

Among them, True Positives (TP) refer to the number of samples that are actually positive and correctly predicted as positive by the model. True Negatives (TN) represent the number of samples that are actually negative and correctly predicted as negative. False Positives (FP) denote the number of samples that are actually negative but incorrectly predicted as positive by the model (commonly referred to as false alarms), whereas False Negatives (FN) indicate the number of samples that are actually positive but misclassified as negative (also known as missed detections). In the visualization of the experimental results, TP, TN, FP, and FN pixels are displayed in white, black, green, and red, respectively, to intuitively highlight the model’s classification behavior. The final evaluation of the model's performance relies on metrics such as OA, KC, Pr, Re, and F1, which are derived from the above statistical quantities. Higher values of these metrics generally indicate better CD performance.

\subsection{Experimental Hyperparameter Settings}
\label{subsec6}
\textit{1) Proposed Model Parameter Settings: }The proposed CHDFFN model is implemented using the PyTorch deep learning framework, and training and testing are conducted on an NVIDIA GTX A100 GPU. During the training phase, the Stochastic Gradient Descent (SGD) optimizer is employed with an initial learning rate of 5e-3, and the model is trained for a total of 100 epochs. The input image patch size is set to 9×9, and the batch size is 32. For training and testing partitioning, 30\% of the pixels are randomly selected from both changed and unchanged regions to form the training set, while the remaining pixels are used for testing. The experimental settings are kept consistent across all four datasets.

\textit{2) Comparison Model Parameter Settings: }According to the original papers, for MHCD\cite{wu2024multitask}, we selected an input patch size of 9×9, a batch size of 64, a training set ratio of 5\%, 300 training epochs, and an initial learning rate of 1e-4. For MsFNet\cite{feng2024msfnet}, the input image patch size was 9×9, and the Adam optimizer was used during training. The initial learning rate was set to 5e-4, the batch size to 32, and the number of training epochs to 200. The DIEFEN\cite{wu2024diefen} algorithm used an input image patch size of 5×5, an initial learning rate of 0.001, and was trained with an SGD optimizer with a decay weight of 5e-3. The training set ratio was 1\%, the batch size was 64, and the total number of epochs was 100. For AIWSEN\cite{wu2025aiwsen}, the input image patch size was 7×7, the initial learning rate was 5e-3, and it was trained using an SGD optimizer with a decay weight of 5e-3, with a decay factor of 0.1 every 35 epochs. The training set ratio was 1\%, the batch size was 64, and the total number of epochs was 100. For MSDFFN\cite{luo2023multiscale}, the input image patch size was 9×9, the initial learning rate was 5e-3, and it was trained using an SGD optimizer with a decay weight of 5e-3, with a decay factor of 0.1 every 35 epochs. The training set ratio was 20\%, the batch size was 32, and the total number of epochs was 100. The DA-Former\cite{wang2023semi} detection architecture used an input image patch size of 9×9, an initial learning rate of 5e-4, and was trained using the Adam optimizer. The training set ratio was uniformly set to 1\%, the batch size was 64, and the total number of epochs was 600.

\subsection{Experimental results}
\label{subsec7}
\textit{1) Experimental Results on the Farmland Dataset: }Table \ref{tab:table1} presents the results of each model on the Farmland dataset. Compared with the AIWSEN and DA-former methods, other methods perform better in all metrics, with significant improvements in OA and KC. This indicates that the overall performance of the AIWSEN and DA-former algorithms is relatively weak, and their ability to identify changed regions is insufficient, leading to unsatisfactory classification results. Compared with AIWSEN, DA-former, and DIEFEN, MHCD, MsFNet, MSDFFN, and the proposed method generally exhibit better performance. This is because these methods can more effectively capture the change features of farmland scenes, thereby improving the accuracy of change detection. After introducing better feature extraction and fusion mechanisms, MSDFFN and the proposed method outperform other methods with relatively average performance. This suggests that efficient mechanisms help the network learn more accurate farmland change features to adapt to different land cover change scenarios. Compared with MSDFFN, the proposed method achieves further improvements in all metrics due to its better balance between precision and recall. Therefore, CHDFFN has better comprehensive performance in the change detection task on the farmland dataset.
\begin{table}[width=.9\linewidth,cols=6]
	\caption{Comparison between CHDFFN and various methods on the Farmland dataset}
	\label{tab:table1}
	\begin{tabular*}{\tblwidth}{@{} LLLLLL@{} }
		\toprule
		Method & OA (\%) & KC (\%) & F1 (\%) & Pr (\%) & Re (\%) \\
		\midrule  
		MHCD & 96.52 & 91.90 & 94.44 & 94.32 & 94.55 \\
		MsFNet & 97.26 & 93.66 & 95.66 & 94.67 & 96.68 \\
		DIEFEN & 93.47 & 85.32 & 90.19 & 85.04 & 96.00 \\
		AIWSEN & 90.13 & 77.72 & 85.06 & 80.71 & 89.90 \\
		MSDFFN & 98.08 & 95.55 & 96.95 & 96.13 & 97.79 \\
		DA-former & 90.19 & 78.14 & 85.49 & 79.48 & 92.48 \\
		Ours & \textbf{98.63} & \textbf{96.87} & \textbf{98.07} & \textbf{97.80} & \textbf{98.34} \\
		\bottomrule  
	\end{tabular*}
\end{table}
\begin{figure}
	\centering
	\includegraphics[width=.9\columnwidth]{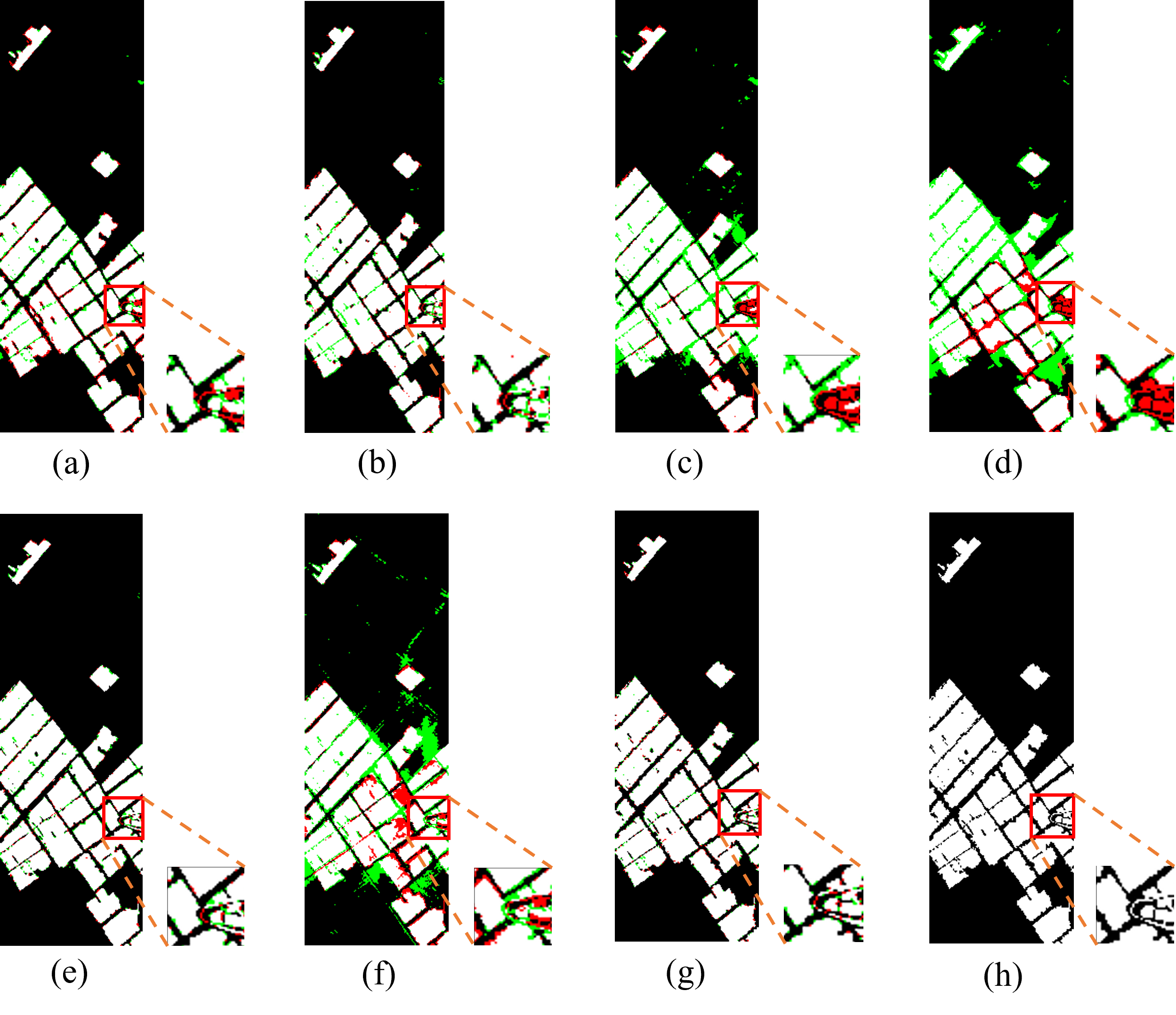}
	\caption{Visualization results of different detection methods on the Farmland dataset. (a) MHCD, (b) MsFNet, (c) DIEFEN, (d) AIWSEN, (e) MSDFFN, (f) DA-former, (g) Ours, (h) Change ground truth.}
	\label{fig_13}
\end{figure}

To present the experimental results more intuitively, we visualized the data in Table \ref{tab:table1} as shown in Fig. \ref{fig_13}. Overall, the result images of DIEFEN, AIWSEN, and DA-former contain a significant number of green pixels, indicating these methods have limited ability to correctly identify unchanged regions. In contrast, MHCD and MsFNet show fewer green and red pixels, suggesting moderate levels of false positives and false negatives. MSDFFN and CHDFFN further reduce such errors, with their result images exhibiting minimal green and red pixels. To illustrate these differences more clearly, we performed local magnification on the visual results, as shown in Fig. \ref{fig_13} (a)-(h). It can be observed that Fig. \ref{fig_13} (a) and Fig. \ref{fig_13} (b) display noticeably more green and red pixels than Fig. \ref{fig_13} (h), while Fig. \ref{fig_13} (e) and Fig. \ref{fig_13} (g) contain similar amounts. A detailed inspection reveals that the locally magnified results of the six baseline methods (DIEFEN, AIWSEN, DA-former, MHCD, MsFNet, and MSDFFN) contain more errors than CHDFFN. Overall, the CHDFFN model produces results with the fewest green and red pixels, indicating lower false positive and false negative rates. This demonstrates the superior performance of the proposed method in accurately detecting change regions.

\textit{2) Experimental Results on the Hermiston2 Dataset: }Table \ref{tab:table2} presents the performance of each model on the Hermiston2 dataset. Compared with the MHCD method, other supervised learning methods achieve better performance across all evaluation metrics, particularly in OA and KC. This suggests that MHCD exhibits inferior classification performance and a weak capacity to distinguish between changed and unchanged regions, thereby limiting its overall effectiveness. In contrast, methods such as MsFNet, AIWSEN, MSDFFN, and the proposed CHDFFN consistently yield superior results relative to MHCD, DIEFEN, and DA-Former, due to their enhanced ability to extract discriminative change features and improve change detection accuracy. Notably, MSDFFN and CHDFFN, which incorporate advanced feature learning mechanisms, outperform those relying solely on conventional feature extraction, underscoring the importance of efficient feature representation in adapting to complex change detection scenarios. Moreover, compared with MSDFFN, our CHDFFN achieves further improvements in overall performance by better balancing precision and recall, leading to more accurate and comprehensive identification of changed pixels.
\begin{table}[width=.9\linewidth,cols=6]
	\caption{Comparison between CHDFFN and various methods on the Hermiston2 dataset}
	\label{tab:table2}
	\begin{tabular*}{\tblwidth}{@{} LLLLLL@{} }
		\toprule  
		Method & OA (\%) & KC (\%) & F1 (\%) & Pr (\%) & Re (\%) \\
		\midrule  
		MHCD     & 87.80 & 62.99 & 70.62 & 77.23 & 65.05 \\
		MsFNet   & 95.30 & 86.02 & 88.99 & 94.30 & 84.25 \\
		DIEFEN   & 93.91 & 82.13 & 86.02 & 89.16 & 83.09 \\
		AIWSEN   & 95.26 & 86.50 & 89.57 & 88.78 & 90.34 \\
		MSDFFN   & 97.66 & 93.26 & 94.76 & 95.74 & 93.81 \\
		DA-former& 94.26 & 83.14 & 86.80 & 90.16 & 83.68 \\
		Ours     & \textbf{98.02} & \textbf{94.30} & \textbf{96.08} & \textbf{96.18} & \textbf{95.99} \\
		\bottomrule  
	\end{tabular*}
\end{table}
\begin{figure}
	\centering
	\includegraphics[width=.9\columnwidth]{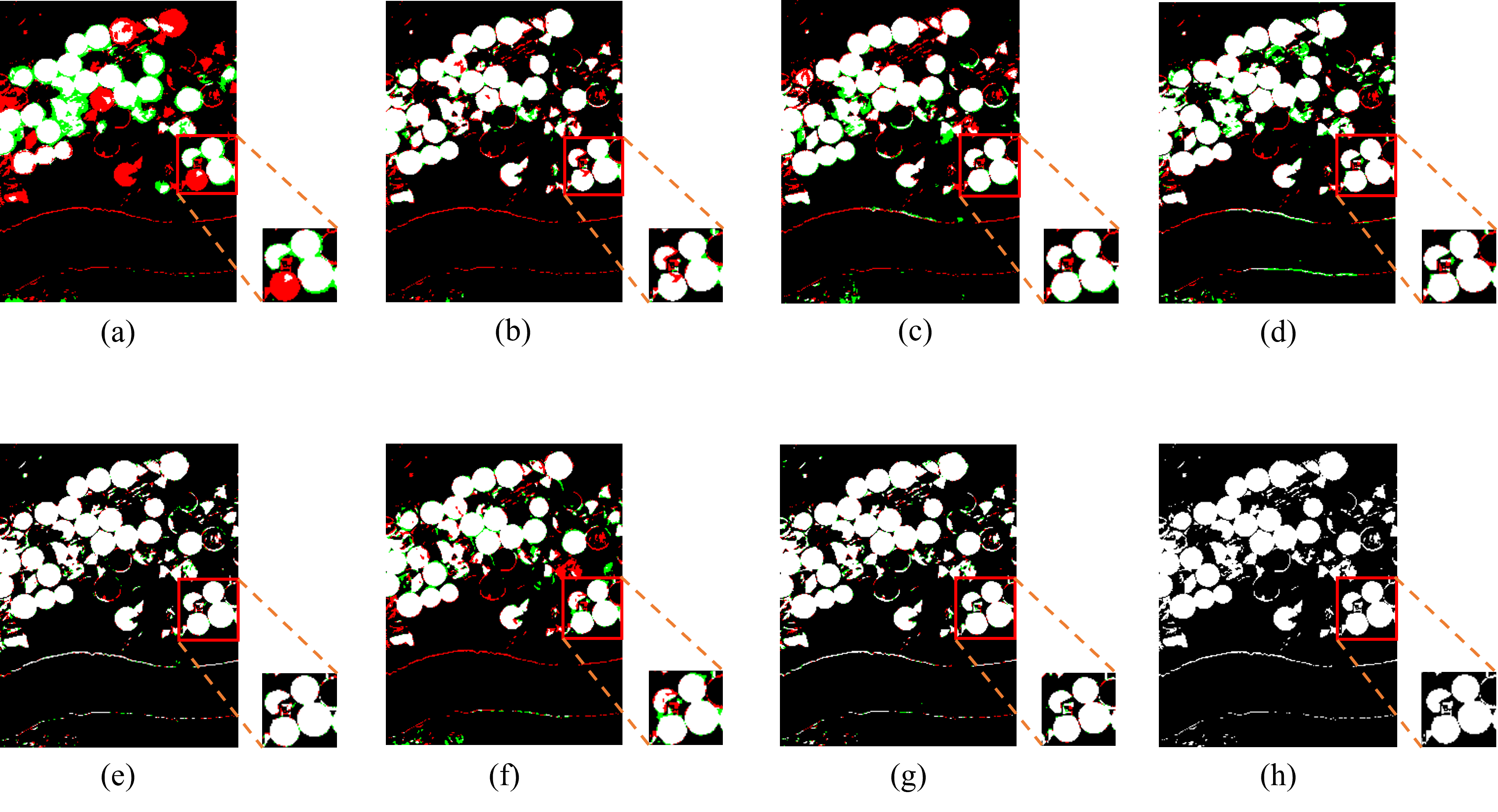}
	\caption{Visualization results of different detection methods on the Hermiston2 dataset. (a) MHCD, (b) MsFNet, (c) DIEFEN, (d) AIWSEN, (e) MSDFFN, (f) DA-former, (g) Ours, (h) Change ground truth.}
	\label{fig_14}
\end{figure}

For ease of intuitive observation, the experimental results in Table \ref{tab:table2} are converted into visual representations, as shown in Fig. \ref{fig_14}. From the overall visualization, the result images generated by MHCD, MsFNet, and DIEFEN contain a large number of red pixels and relatively few green pixels, indicating a high false negative rate in these methods. In contrast, the visual outputs of AIWSEN and DA-Former display only a small number of red and green pixels, suggesting that both methods achieve relatively low false positive and false negative rates. Meanwhile, MSDFFN and CHDFFN produce result images with very few red and green pixels, indicating more accurate detection performance. To further illustrate these observations, localized zoom-in views of the result images are provided in Figs. \ref{fig_14} (a)–(h). From these magnified views, it can be seen that Figs. \ref{fig_14} (a), \ref{fig_14} (b), \ref{fig_14} (c), \ref{fig_14} (d), and \ref{fig_14} (f) contain noticeably more red and green pixels compared to Figs. \ref{fig_14} (e) and \ref{fig_14} (g), while the pixel distribution in Fig. \ref{fig_14} (e) is visually similar to that in Fig. \ref{fig_14} (g). These detailed comparisons reveal that the six methods (DIEFEN, AIWSEN, DA-Former, MHCD, MsFNet, and MSDFFN) generally exhibit more detection errors (false positives and false negatives) than CHDFFN. In summary, the CHDFFN method proposed in this study consistently produces fewer green and red pixels across visual results, reflecting a lower false positive rate and false negative rate. This confirms the superior change detection capability of CHDFFN compared to other baseline methods.

\textit{3) Experimental Results on the Hermiston Dataset: }Table \ref{tab:table3} presents the performance of each model on the Hermiston dataset. Compared to the DIEFEN method, other approaches achieve better results across all evaluation metrics, with particularly significant improvements in Pr and OA. This suggests that the DIEFEN algorithm tends to produce a higher number of false positives and lacks the ability to accurately delineate changed regions, thereby limiting its overall performance. In contrast, methods such as MHCD, AIWSEN, MSDFFN, and the proposed approach demonstrate consistently superior performance compared to DIEFEN and DA-Former. This improvement can be attributed to their enhanced capability in accurately detecting changed areas, effectively suppressing false positives, and improving overall change detection performance. Among these, MSDFFN and the proposed method exhibit the most remarkable results, benefiting from more efficient feature extraction and processing mechanisms. This underscores the importance of advanced architectures in enabling the network to capture finer-grained change features and handle complex land surface changes. Furthermore, compared to MSDFFN, the proposed method achieves additional gains across all metrics. These improvements stem from its superior feature fusion strategy and more effective classification decisions, leading to more robust and comprehensive detection performance.
\begin{table}[width=.9\linewidth,cols=6]
	\caption{Comparison between CHDFFN and various methods on the Hermiston dataset}
	\label{tab:table3}
	\begin{tabular*}{\tblwidth}{@{} LLLLLL@{} }
		\toprule
		Method    & OA (\%) & KC (\%) & F1 (\%) & Pr (\%) & Re (\%) \\
		\midrule
		MHCD      & 99.17   & 96.32 & 96.80   & 96.06   & 97.55   \\
		MsFNet    & 99.11   & 96.00 & 96.51   & 97.40   & 95.63   \\
		DIEFEN    & 98.58   & 93.83 & 94.64   & 91.64   & 97.86   \\
		AIWSEN    & 98.91   & 95.12 & 95.75   & 95.27   & 96.22   \\
		MSDFFN    & 99.55   & 98.01 & 98.26   & 97.91   & \textbf{98.61} \\
		DA-former & 98.73   & 94.30 & 95.03   & 94.92   & 95.13   \\
		Ours      & \textbf{99.55} & \textbf{98.62} & \textbf{98.41} & \textbf{98.29} & 98.52 \\
		\bottomrule
	\end{tabular*}
\end{table}
\begin{figure}
	\centering
	\includegraphics[width=.9\columnwidth]{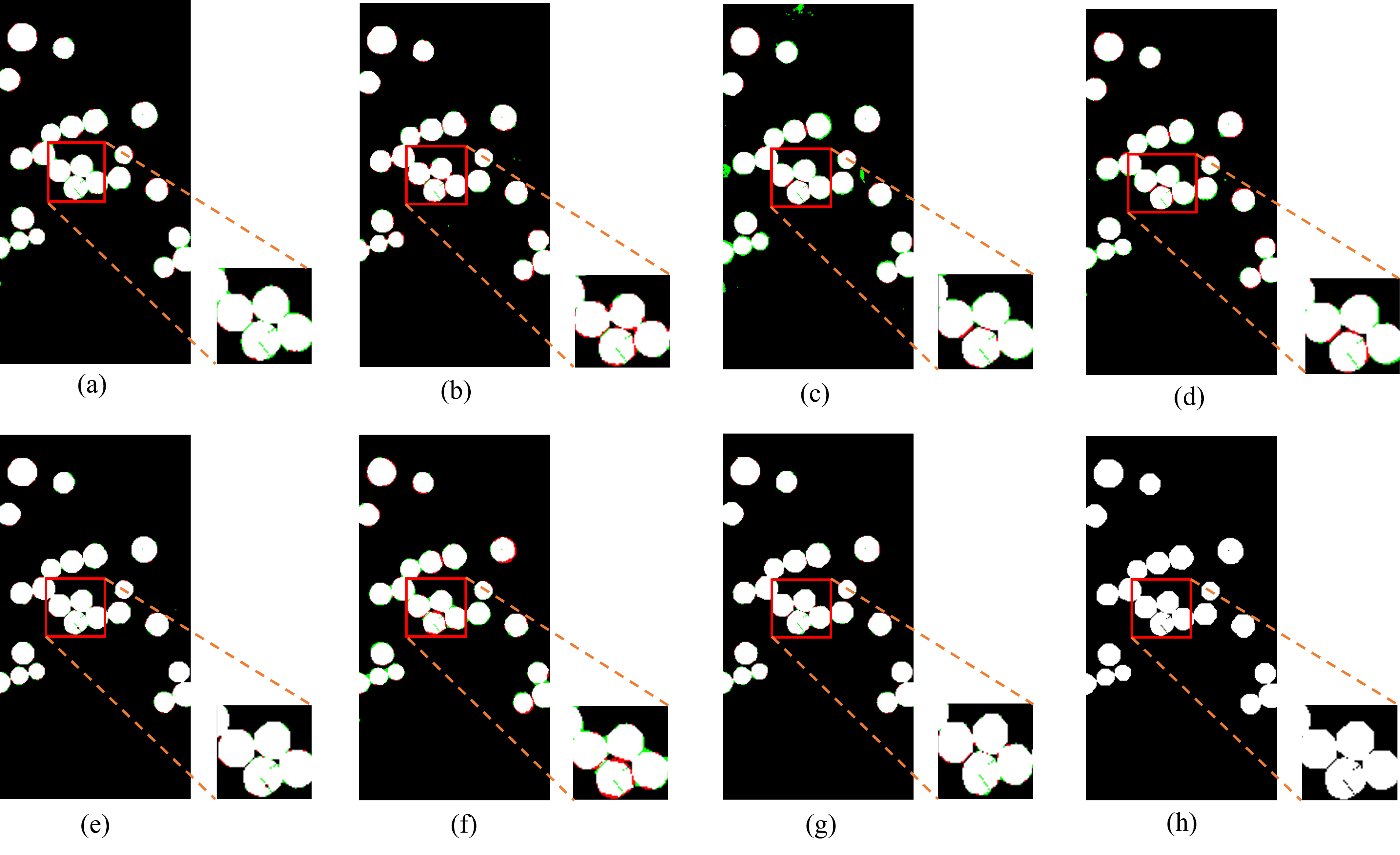}
	\caption{Visualization results of different detection methods on the Hermiston dataset. (a) MHCD, (b) MsFNet, (c) DIEFEN, (d) AIWSEN, (e) MSDFFN, (f) DA-former, (g) Ours, (h) Change ground truth.}
	\label{fig_15}
\end{figure}

For more intuitive observation, the experimental results in Table \ref{tab:table3} are visualized in Fig. \ref{fig_15}. Overall, the DIEFEN results exhibit a marginally higher proportion of green pixels compared to other methods, suggesting a relatively higher false positive rate. In contrast, other methods show a relatively balanced distribution of red and green pixels, indicating similar performance levels. To further clarify this observation, we provide localized enlargements of the visual outputs in Fig. \ref{fig_15} (a)–(h). From these magnified regions, it is evident that Fig. \ref{fig_15} (b) and Fig. \ref{fig_15} (f) contain the highest concentration of red pixels, while Fig. \ref{fig_15} (a) and Fig. \ref{fig_15} (c) show a relatively greater presence of green pixels. Additionally, compared to the locally magnified image of CHDFFN, the corresponding regions of the six other methods (DIEFEN, AIWSEN, DA-Former, MHCD, MsFNet, and MSDFFN) contain more red and green pixels, although the differences are not particularly large. In summary, the CHDFFN model produces visual outputs with fewer red and green pixels, implying lower false positive and false negative rates. This suggests that the proposed method achieves superior change detection performance compared to other baseline methods.

\textit{4) Experimental Results on the River Dataset: }Table \ref{tab:table4} presents the results of various models on the River dataset. Compared with DA-former and MsFNet, the other methods achieve better performance across all evaluation metrics, with particularly notable improvements in OA and KC. This indicates that DA-former and MsFNet do not exhibit significant advantages in classification tasks. Their overall performance is relatively weak, and they demonstrate limited ability in detecting change information. In contrast, MHCD, MSDFFN, and the proposed CHDFFN generally show superior performance compared to DA-former, MsFNet, DIEFEN, and AIWSEN. These methods are more effective in handling the complex scenarios presented in this dataset, thereby improving both the accuracy and coverage of the classification. Among them, MSDFFN and CHDFFN benefit from stronger feature learning and processing capabilities, leading to better overall results than methods with more moderate performance. This suggests that enhanced feature processing enables the network to capture more discriminative change features, which is crucial for accurate change detection in the River dataset. Furthermore, the proposed CHDFFN achieves further improvements over MSDFFN in all metrics. This is attributed to its ability to better balance precision and recall, resulting in more accurate identification of change information and superior overall performance.
\begin{table}[width=.9\linewidth,cols=6]
	\caption{Comparison between CHDFFN and various methods on the River dataset}
	\label{tab:table4}
	\begin{tabular*}{\tblwidth}{@{} LLLLLL@{} }
		\toprule
		Method    & OA (\%) & KC (\%) & F1 (\%) & Pr (\%) & Re (\%) \\
		\midrule
		MHCD      & 96.38   & 75.34 & 77.29   & 85.01   & 70.86   \\
		MsFNet    & 95.47   & 69.18 & 71.62   & 78.65   & 65.75   \\
		DIEFEN    & 95.83   & 70.76 & 72.98   & 83.40   & 64.88   \\
		AIWSEN    & 96.27   & 75.89 & 77.92   & 80.35   & 75.63   \\
		MSDFFN    & 97.96   & 86.96 & 88.07   & 89.47   & 86.72   \\
		DA-former & 95.36   & 69.56 & 72.06   & 76.17   & 68.73   \\
		Ours      & \textbf{98.29} & \textbf{89.39} & \textbf{91.15} & \textbf{90.95} & \textbf{91.35}   \\
		\bottomrule
	\end{tabular*}
\end{table}
\begin{figure}
	\centering
	\includegraphics[width=.9\columnwidth]{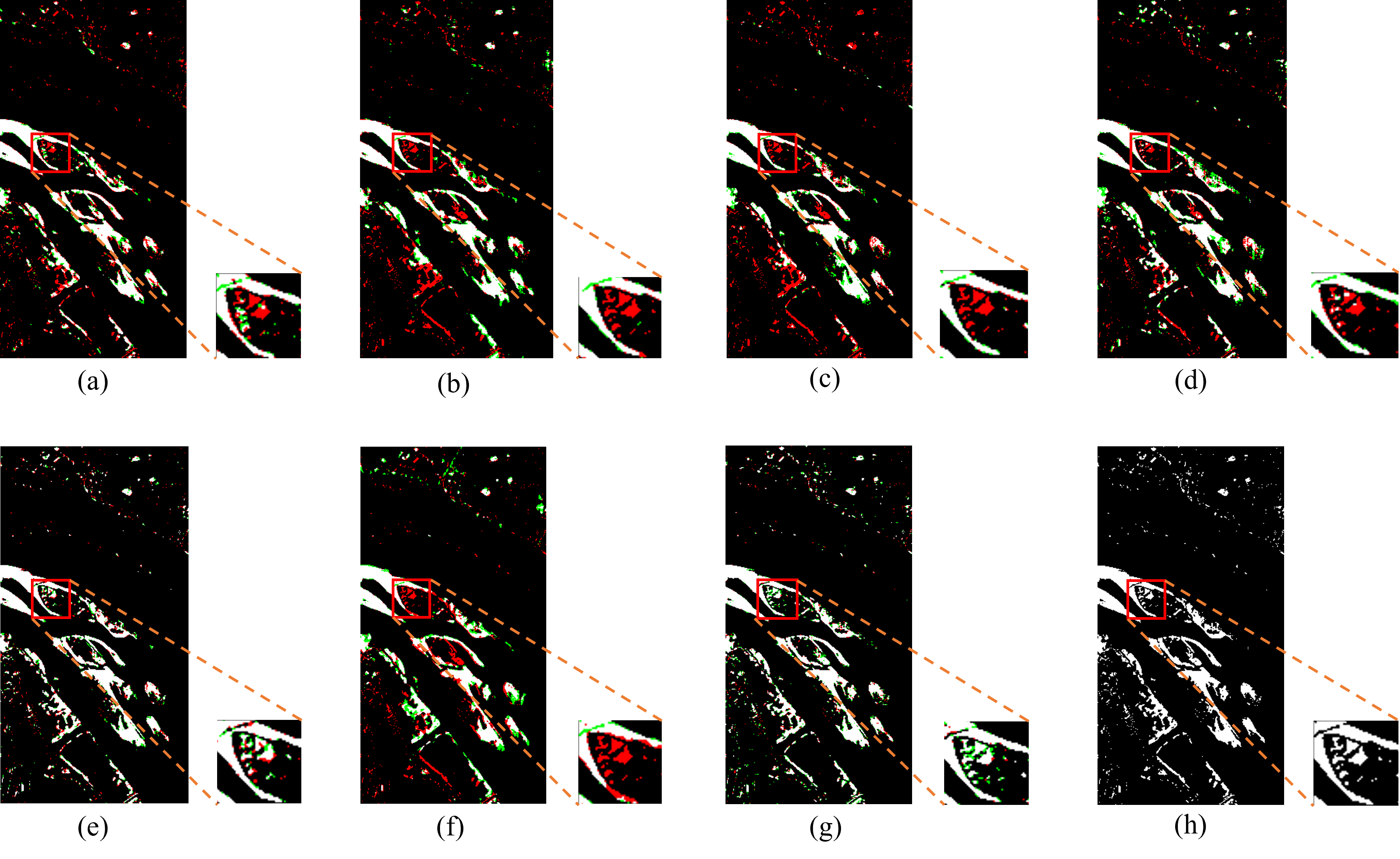}
	\caption{Visualization results of different detection methods on the River dataset. (a) MHCD, (b) MsFNet, (c) DIEFEN, (d) AIWSEN, (e) MSDFFN, (f) DA-former, (g) Ours, (h) Change ground truth.}
	\label{fig_16}
\end{figure}

For intuitive observation, the experimental results in Table \ref{tab:table4} are visualized in Fig. \ref{fig_16}. Overall, the number of red and green pixels in Fig. \ref{fig_16} (a)–(d) and Fig. \ref{fig_16}(f) is significantly higher than in Fig. \ref{fig_16} (e) and Fig. \ref{fig_16} (g). Notably, these images exhibit regions with an excess of red pixels and a scarcity of green pixels, indicating that the MHCD, MsFNet, DIEFEN, AIWSEN, and DA-Former methods suffer from high false negative and false positive rates. In contrast, the result images of MSDFFN and CHDFFN contain relatively fewer red and green pixels, suggesting lower false detection rates. To further support this observation, local magnification was applied to all visualized results. The zoomed-in views show that the number of green pixels in Fig. \ref{fig_16} (e) is slightly greater than in Fig. \ref{fig_16} (g), and the red pixel count in the magnified results of the five aforementioned methods is higher than that of CHDFFN. In comparison, the CHDFFN method proposed in this study produces the fewest red and green pixels overall, indicating its superior performance in change detection with reduced false positives and false negatives.

\subsection{Ablation Experiments}
\label{subsec8}
To more clearly demonstrate the effectiveness of the designed modules, ablation experiments were conducted on each of the four modules across four datasets, including the multiscale convolution encoder module (MSCE), DCCSA module, SSCFL module, and AFAF module. Specifically, four groups of experiments were designed by sequentially removing one module at a time, denoted as A to D. 

The experimental results are shown in Table \ref{tab:table5}. Compared with the complete model, the accuracy of the ablated models generally declines, with the complete model achieving the best overall performance. These results provide strong evidence for the effectiveness of each module. Specifically, for the Farmland and River datasets, the OA, KC, F1, Re, and Pr metrics of the complete model all reach optimal values. On the Hermiston dataset, the highest Re and Pr values are observed in the experiment using the module combination `` MSCE + DCCSA + AFAF '', which surpass those of the complete model. It should be noted that Recall reflects the proportion of correctly identified changed pixels relative to the total number of changed pixels, whereas Precision indicates the proportion of correctly predicted changed pixels among all predicted changes. Due to the imbalance between changed and unchanged pixels, it is possible to observe high Re or Pr values individually, which may result in misleading evaluations if either metric is considered in isolation. As the harmonic mean of Pr and Re, the F1 score balances the two and serves as a more appropriate evaluation metric. Notably, the complete CHDFFN model achieves the highest F1 score across all datasets. Overall, the proposed CHDFFN method effectively identifies changes in ground object states by analyzing the differences between hyperspectral data across different temporal phases. While certain modules may exhibit minor adverse effects on specific datasets, they consistently enhance change detection performance across the majority of experimental scenarios.
\begin{table}
	\caption{Ablation experiment results of module comparison on the four datasets}
	\label{tab:table5}
	\begin{tabular*}{\tblwidth}{@{} LLLLLLL@{} }
		\toprule
		Dataset & Module & A & B & C & D & Full model \\
		\midrule
		& MSCE    & $\times$ & $\checkmark$ & $\checkmark$ & $\checkmark$ & $\checkmark$ \\
		& DCCSA  & $\checkmark$ & $\times$ & $\checkmark$ & $\checkmark$ & $\checkmark$ \\
		& SSCFL  & $\checkmark$ & $\checkmark$ & $\times$ & $\checkmark$ & $\checkmark$ \\
		& AFAF   & $\checkmark$ & $\checkmark$ & $\checkmark$ & $\times$ & $\checkmark$ \\
		\midrule
		\multirow{5}{*}{\centering Farmland} 
		& OA (\%)   & 97.16 & 97.75 & 98.37 & 97.59 & \textbf{98.63} \\
		& KC (\%)   & 95.09 & 96.56 & 96.28 & 95.46 & \textbf{96.87} \\
		& F1 (\%)   & 96.11 & 97.38 & 97.02 & 95.52 & \textbf{98.07} \\
		& Pr (\%)  & 96.14 & 97.42 & 97.08 & 95.77 & \textbf{97.80} \\
		& Re (\%)   & 96.08 & 97.35 & 96.96 & 95.28 & \textbf{98.34} \\
		\midrule
		\multirow{5}{*}{\centering Hermiston2} 
		& OA (\%)   & 96.86 & 97.86 & 97.48 & 97.37 & \textbf{98.02} \\
		& KC (\%)   & 93.80 & 93.81 & 93.34 & 94.46 & \textbf{94.30} \\
		& F1 (\%)   & 93.83 & 95.10 & 94.50 & 94.77 & \textbf{96.08} \\
		& Pr (\%)  & 94.05 & 94.95 & 96.08 & 95.03 & \textbf{96.18} \\
		& Re (\%)   & 93.61 & 95.26 & 92.98 & 94.51 & \textbf{95.99} \\
		\midrule
		\multirow{5}{*}{\centering Hermiston} 
		& OA (\%)   & 97.73 & 98.75 & 98.21 & 98.41 & \textbf{99.55} \\
		& KC (\%)   & 96.97 & 98.51 & 98.33 & 98.18 & \textbf{98.62} \\
		& F1 (\%)   & 97.03 & 97.83 & 98.08 & 97.37 & \textbf{98.41} \\
		& Pr (\%)  & 97.05 & 97.66 & 98.16 & 98.06 & \textbf{98.29} \\
		& Re (\%)   & 97.02 & 98.01 & 98.01 & 96.69 & \textbf{98.52} \\
		\midrule
		\multirow{5}{*}{\centering River} 
		& OA (\%)   & 97.02 & 98.11 & 97.79 & 97.87 & \textbf{98.29} \\
		& KC (\%)   & 87.15 & 89.93 & 87.88 & 89.44 & \textbf{89.39} \\
		& F1 (\%)   & 86.26 & 89.79 & 88.78 & 89.79 & \textbf{91.15} \\
		& Pr (\%)  & 85.10 & 89.96 & 88.39 & 89.71 & \textbf{90.95} \\
		& Re (\%)   & 87.46 & 89.62 & 89.17 & 89.87 & \textbf{91.35} \\
		\bottomrule
	\end{tabular*}
\end{table}

\subsection{Discussion}
\label{subsec9}
\textit{1) Discussion on the setting of training sample proportion: }As a key factor in deep learning model training, the training sample proportion significantly affects both feature learning and generalization of the CHDFFN model. To verify the suitability of setting it to 30\%, comparative experiments were conducted on four datasets with proportions of 5\%, 10\%, 20\%, and 30\%, while keeping all other hyperparameters fixed. The results in Table \ref{tab:table6} support the comparative analysis and validation of this choice.

The results show that a 30\% training sample proportion yields peak OA and KC on the Farmland, Hermiston, and River datasets, confirming its suitability for feature learning. Although the Hermiston2 dataset achieves a local optimum in Pr at 20\%, likely due to the higher proportion of unchanged pixels at 30\%, the 30\% setting offers the most balanced performance overall, mitigating underfitting at small proportions and avoiding overfitting at larger ones.
\begin{table}
	\caption{Results of different training sample proportions on the four datasets}
	\label{tab:table6}
	\begin{tabular*}{\tblwidth}{@{} LLLLLL@{} }
		\toprule
		Dataset & Sample ratio (\%) & 5 & 10 & 20 & 30 \\
		\midrule
		\multirow{5}{*}{Farmland} 
		& OA (\%) & 96.22 & 97.07 & 98.11 & \textbf{98.63} \\
		& KC (\%) & 92.33 & 93.96 & 95.56 & \textbf{96.87} \\
		& F1 (\%) & 95.09 & 95.75 & 96.76 & \textbf{98.07} \\
		& Pr (\%) & 94.09 & 95.17 & 96.22 & \textbf{97.80} \\
		& Re (\%) & 96.11 & 96.33 & 97.31 & \textbf{98.34} \\
		\midrule
		\multirow{5}{*}{Hermiston2} 
		& OA (\%) & 95.69 & 96.64 & 97.75 & \textbf{98.02} \\
		& KC (\%) & 88.99 & 90.27 & 93.59 & \textbf{94.30} \\
		& F1 (\%) & 90.64 & 92.17 & 94.89 & \textbf{96.08} \\
		& Pr (\%) & 90.49 & 93.08 & 96.16 & \textbf{96.18} \\
		& Re (\%) & 90.79 & 91.28 & 93.65 & \textbf{95.99} \\
		\midrule
		\multirow{5}{*}{Hermiston} 
		& OA (\%) & 98.09 & 98.85 & 99.12 & \textbf{99.55} \\
		& KC (\%) & 96.29 & 97.61 & 98.15 & \textbf{98.62} \\
		& F1 (\%) & 96.54 & 96.73 & 97.61 & \textbf{98.41} \\
		& Pr (\%) & 95.96 & 96.20 & 97.29 & \textbf{98.29} \\
		& Re (\%) & 97.13 & 97.27 & 97.94 & \textbf{98.52} \\
		\midrule
		\multirow{5}{*}{River} 
		& OA (\%) & 95.52 & 96.80 & 97.73 & \textbf{98.29} \\
		& KC (\%) & 79.66 & 84.15 & 87.65 & \textbf{89.39} \\
		& F1 (\%) & 80.90 & 85.86 & 88.67 & \textbf{91.15} \\
		& Pr (\%) & 82.58 & 85.07 & 89.29 & \textbf{90.95} \\
		& Re (\%) & 79.28 & 86.67 & 88.06 & \textbf{91.35} \\
		\bottomrule
	\end{tabular*}
\end{table}

\textit{2) Discussion on the size of input image patches: }The input image patch size critically affects hyperspectral change detection performance. To assess the impact of spatial context, multi-scale experiments were conducted on four datasets with patch sizes of 5×5, 7×7, and 9×9, while keeping all other hyperparameters fixed. The results are summarized in Table \ref{tab:table7} for further analysis.

The experimental results show that a 9×9 input patch size consistently achieves the best OA and KC across all datasets. This is attributed to its ability to capture richer neighborhood context, enlarge the receptive field, and enhance spatial–spectral feature fusion, which is crucial for distinguishing subtle changes in complex scenarios. By providing more discriminative contextual information, the 9×9 setting improves feature robustness while maintaining computational efficiency, and is therefore adopted as the final configuration.
\begin{table}
	\caption{Results of different input image patch sizes on the four datasets}
	\label{tab:table7}
	\begin{tabular*}{\tblwidth}{@{} LLLLL@{} }
		\toprule
		Dataset & Patch size & $5\times5$ & $7\times7$ & $9\times9$ \\
		\midrule
		\multirow{5}{*}{Farmland} 
		& OA (\%) & 96.89 & 98.01 & \textbf{98.63} \\
		& KC (\%) & 95.33 & 96.58 & \textbf{96.87} \\
		& F1 (\%) & 95.76 & 96.80 & \textbf{98.07} \\
		& Pr (\%) & 96.16 & 96.21 & \textbf{97.80} \\
		& Re (\%) & 95.36 & 97.39 & \textbf{98.34} \\
		\midrule
		\multirow{5}{*}{Hermiston2} 
		& OA (\%) & 97.02 & 97.96 & \textbf{98.02} \\
		& KC (\%) & 93.76 & 93.98 & \textbf{94.30} \\
		& F1 (\%) & 94.93 & 95.17 & \textbf{96.08} \\
		& Pr (\%) & 94.81 & 94.68 & \textbf{96.18} \\
		& Re (\%) & 95.05 & 95.66 & \textbf{95.99} \\
		\midrule
		\multirow{5}{*}{Hermiston} 
		& OA (\%) & 97.98 & 98.99 & \textbf{99.55} \\
		& KC (\%) & 98.32 & 98.64 & \textbf{98.62} \\
		& F1 (\%) & 97.46 & 97.87 & \textbf{98.41} \\
		& Pr (\%) & 98.02 & 97.48 & \textbf{98.29} \\
		& Re (\%) & 96.91 & 98.27 & \textbf{98.52} \\
		\midrule
		\multirow{5}{*}{River} 
		& OA (\%) & 97.38 & 97.83 & \textbf{98.29} \\
		& KC (\%) & 88.04 & 88.81 & \textbf{89.39} \\
		& F1 (\%) & 88.07 & 88.65 & \textbf{91.15} \\
		& Pr (\%) & 86.98 & 87.72 & \textbf{90.95} \\
		& Re (\%) & 89.19 & 89.60 & \textbf{91.35} \\
		\bottomrule
	\end{tabular*}
\end{table}

\textit{3) Discussion on input batch size: }As a key hyperparameter in deep learning, batch size influences both performance and efficiency in HCD tasks. To determine the optimal setting, this study compared batch sizes of 32, 64, and 128 under identical conditions, with results summarized in Table \ref{tab:table8}.

Analysis of the four datasets shows that larger batch sizes lead to a consistent decline in OA and KC. A batch size of 32 yields optimal results across most metrics, with only slightly lower Pr on the Hermiston dataset due to class imbalance. Overall, smaller batch sizes enhance performance, and thus 32 is adopted as the final setting, balancing efficiency and accuracy in HCD tasks.
\begin{table}
	\caption{Results of different input batch size on the four datasets}
	\label{tab:table8}
	\begin{tabular*}{\tblwidth}{@{} LLLLL@{} }
		\toprule
		Dataset & Batch size & 32 & 64 & 128 \\
		\midrule
		\multirow{5}{*}{Farmland} 
		& OA (\%) & \textbf{98.63} & 97.89 & 97.01 \\
		& KC (\%) & \textbf{96.87} & 96.02 & 95.18 \\
		& F1 (\%) & \textbf{98.07} & 97.24 & 96.50 \\
		& Pr (\%) & \textbf{97.80} & 96.67 & 96.04 \\
		& Re (\%) & \textbf{98.34} & 97.82 & 96.96 \\
		\midrule
		\multirow{5}{*}{Hermiston2} 
		& OA (\%) & \textbf{98.02} & 97.15 & 96.80 \\
		& KC (\%) & \textbf{94.30} & 93.09 & 92.37 \\
		& F1 (\%) & \textbf{96.08} & 95.78 & 94.75 \\
		& Pr (\%) & 96.18 & \textbf{96.68} & 95.88 \\
		& Re (\%) & \textbf{95.99} & 94.89 & 93.64 \\
		\midrule
		\multirow{5}{*}{Hermiston} 
		& OA (\%) & \textbf{99.55} & 98.54 & 98.12 \\
		& KC (\%) & \textbf{98.62} & 97.55 & 96.82 \\
		& F1 (\%) & \textbf{98.41} & 97.59 & 96.64 \\
		& Pr (\%) & \textbf{98.29} & 97.66 & 97.25 \\
		& Re (\%) & \textbf{98.52} & 97.53 & 96.03 \\
		\midrule
		\multirow{5}{*}{River} 
		& OA (\%) & \textbf{98.29} & 97.17 & 96.02 \\
		& KC (\%) & \textbf{89.39} & 87.38 & 86.72 \\
		& F1 (\%) & \textbf{91.15} & 89.23 & 87.96 \\
		& Pr (\%) & \textbf{90.95} & 88.89 & 87.28 \\
		& Re (\%) & \textbf{91.35} & 89.58 & 88.66 \\
		\bottomrule
	\end{tabular*}
\end{table}

\section{Conclusion}
\label{sec4}
To address the insufficient exploitation of multiscale features in existing HCD methods, this paper proposes a detection framework referred to as CHDFFN, which incorporates several carefully designed components, including multiscale encoder-decoder, residual connection, the improved attention mechanism DCCSA, and CNNs. The multiscale feature extraction sub-network is designed to extract and preliminarily fuse features through multiscale convolution blocks and an encoder-decoder structure. Subsequently, the SSCFL module is embedded to more effectively learn and represent spatial-spectral feature. Finally, the proposed AFAF module performs the ultimate fusion of high-level features output by the SSCFL module, followed by an MLP classifier for pixel-level change detection. Extensive experiments conducted on four publicly available datasets demonstrate that CHDFFN can fully exploit hierarchical representations from multi-temporal HSI and multiscale spatial-spectral feature, thereby significantly enhancing the accuracy of hyperspectral change detection. In future work, we plan to develop a semi-supervised change detection framework that leverages a limited amount of labeled data together with abundant unlabeled data. This approach aims to enhance model generalization by mining latent information from the unlabeled data, which is particularly valuable in scenarios where labeled samples are expensive or difficult to obtain.

\newtheorem{theorem}{Theorem}
\printcredits

\section*{Declaration of competing interest}
The authors declare that there are no known competing financial interests or personal relationships that could have influenced the research work reported in this paper.

\section*{Acknowledgments}
This work was supported by the Key Research Project of Hainan Province under Grant ZDYF2021SHFZ093, the Natural Science Foundation of China under Grants 62063004 and 62162022, the Hainan Provincial Natural Science Foundation of China under Grants 2019RC018, 5210N206 and 6190N249, the Major Scientific Project of Zhejiang Lab 2020ND8AD01, and the Scientific Research Foundation for Hainan University (No. KYOD(ZR)21013).

\section*{Data availability}
Data will be made available on request.

\bibliographystyle{cas-model2-names}
\bibliography{ref}

%
%

\end{document}